\documentclass{article}

\usepackage{PRIMEarxiv}

\usepackage{amsmath,amsfonts}
\usepackage{algorithmic}
\usepackage{algorithm}
\usepackage{array}
\usepackage[caption=false,font=normalsize,labelfont=sf,textfont=sf]{subfig}
\usepackage{textcomp}

\usepackage{booktabs}
\usepackage{verbatim}
\usepackage{graphicx}
\usepackage{cite}
\usepackage{array}
\usepackage{xcolor}

\usepackage{multirow}
\usepackage{enumitem}
\usepackage{makecell}
\usepackage{tcolorbox}
\tcbuselibrary{breakable}
\usepackage{color}
\usepackage{array}
\newcolumntype{L}[1]{>{\raggedright\let\newline\\\arraybackslash\hspace{0pt}}m{#1}}
\newcolumntype{C}[1]{>{\centering\let\newline\\\arraybackslash\hspace{0pt}}m{#1}}
\newcolumntype{R}[1]{>{\raggedleft\let\newline\\\arraybackslash\hspace{0pt}}m{#1}}
\usepackage{soul}

\newcommand{\tabitem}{~\llap{\textbullet}~}

\begin{document}

\newenvironment{idea}{\color{black}{}}

\pagestyle{fancy}
\thispagestyle{empty}
\rhead{ \textit{ }} 

\title{Evolutionary Computation for the Design and Enrichment of General-Purpose Artificial \\ Intelligence Systems: Survey and Prospects}

\author{Javier Poyatos$^{1}$, Javier Del Ser$^{2,3}$, Salvador García$^{1}$, Hisao Ishibuchi$^{5}$, Daniel Molina$^{1}$, \\ \textbf{Isaac Triguero}$^{1,4,*}$, \textbf{Bing Xue}$^{6}$, \textbf{Xin Yao}$^{7,8}$, \textbf{Francisco Herrera}$^{1}$ \\
  $^1$ Department of Computer Science and Artificial Intelligence, Andalusian Research Institute \\ in Data Science and Computational Intelligence (DaSCI), University of Granada, Granada, 18071, Spain \\
  $^2$ TECNALIA, Basque Research Technology Alliance (BRTA), Derio, 48160, Spain\\  
  $^3$ University of the Basque Country (UPV/EHU), Bilbao, 48013, Spain \\
  $^4$ School of Computer Science, University of Nottingham, Nottingham, NG8 1BB, United Kingdom\\
  $^5$ Southern University of Science and Technology, 518055 Shen Zhen, China \\
  $^6$ Victoria University of Wellington, School of Engineering and Computer Science, 6012 Wellington, New Zealand \\
  $^7$ CERCIA, School of Computer Science, University of Birmingham, B15 2SQ Birmingham, U.K \\
  $^8$ Department of Computing and Decision Sciences, Lingnan University, Hong Kong, SAR, China \\
  \texttt{$^*$Corresponding author: triguero@decsai.ugr.es} \\
}

\maketitle

\begin{abstract}

In Artificial Intelligence, there is an increasing demand for adaptive models capable of dealing with a diverse spectrum of learning tasks, surpassing the limitations of systems devised to cope with a single task. The recent emergence of General-Purpose Artificial Intelligence Systems (GPAIS) poses model configuration and adaptability challenges at far greater complexity scales than the optimal design of traditional Machine Learning models. Evolutionary Computation (EC) has been a useful tool for both the design and optimization of Machine Learning models, endowing them with the capability to configure and/or adapt themselves to the task under consideration. Therefore, their application to GPAIS is a natural choice. This paper aims to analyze the role of EC in the field of GPAIS, exploring the use of EC for their design or enrichment. We also match GPAIS properties to Machine Learning areas in which EC has had a notable contribution, highlighting recent milestones of EC for GPAIS. Furthermore, we discuss the challenges of harnessing the benefits of EC for GPAIS, presenting different strategies to both design and improve GPAIS with EC, covering tangential areas, identifying research niches, and outlining potential research directions for EC and GPAIS.
\end{abstract}

\keywords{General-purpose AI \and Evolutionary Computation \and Evolutionary Deep Learning \and Auto-ML\and Neuroevolution \and Open-ended evolution}


\section{Introduction}
\newtheorem{definition}{Definition} 


Machine Learning (ML) is a subfield of Artificial Intelligence (AI) that focuses on developing models capable of learning patterns from data. The optimization of these models has been a prominent area of research, yielding significant results \cite{Song2019} by tailoring their structural design and/or hyper-parameters based on several objectives, such as performance, complexity, or robustness, among others \cite{Yang2020,Yao2023robust}. The diversity of optimization goals considered to date in the optimization of ML models reflects the capability of these algorithms to tackle different criteria.

Recently, advances in several areas of ML research such as Deep Learning (DL) \cite{Goodfellow2016} and Large Language Models \cite{Mikolov2013, Hirschberg2015} -- with chatbots of unprecedented performance like ChatGPT \cite{Vandis2023} and models trained to generate code to improve the effectiveness of mutation operators in Evolutionary Computation (EC) \cite{Lehman2024} -- indicate a notable shift towards more generalized AI systems. The key idea of \emph{general-purpose} AI systems (GPAIS) is their unique capability to execute several, potentially diverse modeling problems, with potentials to expand to tasks for which they were not originally designed. This ability to model several tasks and generalize beyond known problems has been highlighted in recent definitions of GPAIS (see \cite{Triguero2024} and references therein discussed). As a result, GPAIS have gained significance in the last year due to their flexibility and adaptability across a wide range of applications. The possibility that GPAIS develop emerging generalization capabilities, though still unproven, has drawn much attention for their practical implications in terms of AI safety \cite{anderljung2023frontier,bengio2024managing}.


As stated previously, multiple approaches to optimize ML models have shown significant competence across various ML domains \cite{Sun2020}. However, recent studies signal a new research trend focusing on the optimization of GPAIS, moving beyond traditional ML \cite{Gutierrez2023,Uuk2023,Campos2023,Barrett2023,Triguero2024}. Areas of interest in optimization such as open-ended evolution \cite{Standish2003,Taylor2016,Packard2019,Zhang2023omni}, quality-diversity optimization \cite{Pugh2017,Cully2018,Bradley2023qd} and novelty search \cite{Lehman2008} have become crucial for GPAIS. In these optimization areas, GPAIS stand at the forefront of the current research. GPAIS exhibits compelling characteristics such as generating new knowledge, adapting to diverse environments, and seamlessly integrating new tasks without performance degradation. Within GPAIS, as in other complex systems, critical decisions shape their design and optimization, addressing challenges like search space dimensionality, evolving objectives, and the need for continuous adaptation. This dynamic landscape offers opportunities to integrate GPAIS with effective optimization techniques.

In this regard, EC stands out as a versatile family of algorithms that have spearheaded the design and optimization of ML models \cite{Harith2019,Telikani2021}, making a substantial impact on both scenarios \cite{Taner2022,Linan2023}. These algorithms have a great impact on other aspects of ML like learning rules \cite{Yao1999}. EC has played a pivotal role in various research areas, yielding valuable contributions to the development of high-quality ML models. A notable example of its success is in the area of Large Language Models merging them with Evolutionary Algorithms (EAs) \cite{Akiba2024}.

EC has shown its capability to evolve programs, solve dynamic optimization problems, or balance several conflicting objectives, among other optimization scenarios. With its history of successes at optimizing narrow AI systems, EC is particularly attractive for the optimization of GPAIS and for tackling the greater challenges it poses compared to narrow AI systems. Coincidentally, certain research areas of EC match some of the core properties of GPAIS, including adaptability to changing problems over time (evolutionary dynamic optimization) or the confluence of multiple objectives in multitask settings (multitask and multi-objective optimization). Additional strengths of EC for GPAIS we may highlight include: EC is domain-agnostic but can incorporate it if available, and it is free from assumptions about data or model properties. EC can simultaneously build model structures and optimize parameters, allowing for creative method development, and is particularly suitable for multi-objective optimization due to its population-based search approach.

This paper analyzes the potential of AI-powered AI that consists of AI models enhanced or even designed by another AI model. When using EC algorithms as an additional layer of abstraction for the design or enhance AI models, we call it EC-powered AI, and we will refer to those models as EC-GPAIS.  Besides introducing the concept and exposing the benefits of EC when applied to design or enrich GPAIS, this position paper also surveys recent contributions falling within EC-GPAIS, ending up with a prospect of the research directions that can drive future efforts in this area. In particular, the objectives of this work are:

\begin{itemize}[leftmargin=*]
    \item \textit{To match the potential of EC with GPAIS}, with the aim of fostering more research in this trending area in AI. To do this, we will carry out the following three objectives:
    
    \begin{itemize}
        \item To study the design and enhancement of GPAIS using EC with a taxonomy of EC-powered AI. Each category of the taxonomy serves as a guideline on how EC can design or improve AI.
        \item To match GPAIS properties with specific ML areas in which EC has made significant contributions, to show that the connection between EC and GPAIS is well-defined, as EC may enable these properties.
        \item To show the recent milestones of EC-GPAIS. Under the definition of GPAIS presented in the next section, we illustrate various works that have contributed to the beginning and progress of this field in recent years, showcasing EC-GPAIS' potential.
    \end{itemize}

    \item \textit{To discuss the challenges of harnessing the EC-GPAIS benefits and strategies to address advances}, focusing on future challenges. To accomplish this, we break it down into the following two sub-objectives: 

        \begin{itemize}
            \item To study the challenges to obtain the benefits of EC in the GPAIS context. Both designing and enriching GPAIS using EC are complex tasks, and we must ensure that we exploit the benefit of their synergy.
            \item To explore strategies that can be implemented with EC to both design and improve GPAIS. These strategies will serve as a guide for present and future developments of GPAIS using EC.
        \end{itemize}

\end{itemize}

The rest of the paper is organized as follows: Section \ref{sec:evocompgpais} covers the background on EC and ML, its relation with GPAIS, and the paradigm of AI-powered AI for GPAIS. Section \ref{sec:eapowerai} examines different options to use EC to design and enhance GPAIS, matches properties of GPAIS with ML areas that have benefited from EC, and shows practical cases as milestones in EC-GPAIS developments. After that, Section \ref{sec:challenges} emphasizes the challenges for harnessing the benefits of EC-GPAIS and also the strategies for addressing GPAIS using EC, along with suitable research areas within EC for implementing these strategies. Finally, Section \ref{sec:conclusions} concludes the paper with a summary of the contributions and main remarks drawn from this study.

\section{Evolutionary Computation and GPAIS} \label{sec:evocompgpais}

This section briefly recalls the importance of EC in ML over time and in the new era of model generation in AI, in Subsection \ref{sec:evotime}. Then, it explains several characteristics of GPAIS in Subsection \ref{sec:gpaisdefprop}. Finally, it shows the relevance of the paradigm of AI-powered AI concerning GPAIS in Subsection \ref{sec:aipowerai}.

\subsection{Evolutionary Computation in Machine Learning} \label{sec:evotime}

EC has provided intelligent mechanisms capable of optimizing models in various environments. As stated by Li et al. \cite{Linan2023}, who reviewed more than five hundred proposals, these algorithms have been widely used to optimize different stages of ML pipelines, including pre-processing, post-processing, and modeling itself. It is worth noting that within the entire pipeline, evolutionary computation has achieved significant importance in specific branches, such as feature selection \cite{XueBing2016}, feature extraction \cite{Mauceri2021}, ensemble modeling \cite{heywood2023evolutionary}, and even in enhancing model design alongside other techniques such as Support Vector Machines \cite{Fu2019} and Decision Trees \cite{Barros2012}.

The rapid growth of DL in recent years has expanded the application boundaries of this field. Diverse evolutionary proposals have facilitated the optimization of the weights, architecture, and configuration of DL models \cite{DelSer2019,Zhan2022}. EC has significantly contributed to progress in every new topic in ML, owing to the development of algorithms that advance knowledge in these areas. These algorithms have been widely studied due to their applicability across a range of domains \cite{Gen2023}. Therefore, in the modern era of AI, where increasingly complex problems arise, the background in evolutionary computation suggests it is poised to become one of the primary mechanisms for the new generation of AI models. 
    
\subsection{GPAIS: Definitions and Properties} \label{sec:gpaisdefprop}

The study of GPAIS departs from a clear distinction from narrow AI systems, which are instead designed for specific tasks. This differentiation has sparked discussions about the characteristics of an AI system to be classed as GPAIS \cite{Uuk2023}. A recent study (\cite{Triguero2024}) provides a comprehensive analysis of GPAIS, posing a clear landmark definition, a characterization, and a classification of these systems through the use of a taxonomy that distinguishes several strategies to build GPAIS. In particular, GPAIS are defined according to the process followed to incorporate a new task into the system, either by retraining the whole system from scratch with that task or by performing an adaptation to solve it while retaining the knowledge captured by the system from previous data. According to \cite{Triguero2024}:

\begin{tcolorbox}[breakable,notitle,boxrule=0pt,colback=gray!10,colframe=gray!20]
\emph{Definition 1 (GPAIS):} A General Purpose Artificial Intelligence System (GPAIS) refers to an advanced AI system capable of effectively performing a range of distinct tasks. Its degree of autonomy and ability is determined by several key characteristics, including the capacity to adapt or perform well on new tasks that arise at a future time, the demonstration of competence in domains for which it was not intentionally and specifically trained, the ability to learn from limited data, and the proactive acknowledgment of its own limitations in order to enhance its performance.
\end{tcolorbox}

Definition 1 \cite{Triguero2024} provides a comprehensive description of the characteristics that GPAIS may possess. In what follows, and based on \cite{Triguero2024}, we briefly present the concepts in GPAIS that will help understand the role that EAs may undertake. 

The key distinguishing feature of GPAIS vs classical narrow AI lies in the ability to simultaneously tackle multiple learning tasks (whether known or unknown). Taking into consideration what we know about those tasks, Triguero et al. \cite{Triguero2024} differentiate between two types of GPAIS:  

\begin{itemize}[leftmargin=*]
    \item \textit{\textbf{Closed-world GPAIS}}: It assumes we have data for a given number of tasks, and those will be the only tasks that will be dealt with. 
    
    \item \textit{\textbf{Open-world GPAIS}}: These systems acknowledge the fact that new tasks may arise and limited (or none) data may be available.     
\end{itemize}

\begin{figure}[h!]
    \centering
    \includegraphics[width=0.7\columnwidth]{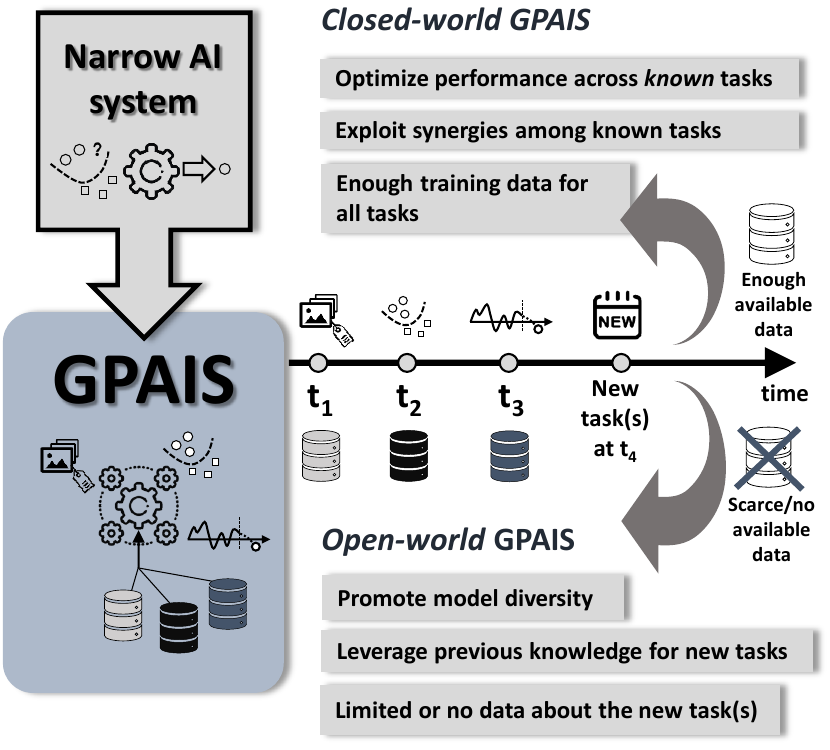}
    \caption{Closed-world vs. Open-world GPAIS (adapted from \cite{Triguero2024}).}
    \label{fig:gpais}
\end{figure}

Figure \ref{fig:gpais} represents the differences between these two settings, highlighting some of the most relevant expectations for them. For example, in the open-world setting, we expect very little data, and leveraging previous knowledge from other tasks becomes imperative. In doing so, open-world GPAIS focus on model diversity/generalization rather than on tuning the model configuration for a particular task, so that they can be used in different problems even in the absence of enough data for the new tasks. The well-known field of Generative AI (GenAI) is composed of systems that are a prime example of open-world GPAIS systems, which are characterized by their autonomy, adaptability to new tasks, competence in domains not intentionally trained for, ability to learn from limited data, and proactive acknowledgment of their own limitations. These features have undergone a complete transformation of AI, as reflected in \cite{RistoMik2024}. 

This definition allows us to distinguish between various degrees of autonomy for GPAIS, however, there are multiple research trends to realize these systems. In the specialized literature, we find two distinct, but not exclusive, approaches. On the one hand, we may attempt to create a single multitask learning model that is general/diverse enough. To do so, the model may require to be trained on broad data (e.g. using self-supervision) to later be adapted (fine-tuned) to specific new tasks. That is the underlying idea of foundation models \cite{Bommasani2021}. On the other hand, we may opt for adding up a new layer of abstraction that would use another AI to help the underlying AI to be more general. This is where EAs may make a difference. The next subsection focuses on describing the different research trends in which an AI model can help another based on the taxonomy proposed in \cite{Triguero2024}.

\subsection{AI-powered AI for GPAIS} \label{sec:aipowerai}

The concept of AI models enhanced or even designed by another AI model is typically known as AI-powered AI. Following the taxonomy proposed in \cite{Triguero2024}, we show a non-exhaustive list of topics in which an AI may be useful to either design or enrich another AI. In terms of designing a general AI model, we may aim at different levels: 

\begin{itemize}[leftmargin=*]
    \item \textbf{Hyper-parameter optimization}: While the use of EC for hyper-parameter optimization is not new or uncommon in narrow AI \cite{Yu2020,Back2023}, here we focus on the idea of tuning a set of hyper-parameters to solve a range of tasks \cite{Huang2020}. In the open world, the challenge lies in finding the best configuration when a new task, potentially with little or no data, arises, exploiting prior knowledge to generalize.
        
    \item \textbf{Automated algorithm selection}: The previous approach can be further improved by determining both the most appropriate AI algorithm and its hyper-parameters. This would enable a more general AI solution that automatically decides the AI technique to use for one or multiple problems at once. In ML, this is typically referred to as AutoML \cite{Hutter2019}. In this field, EAs have made a good contribution, aiming to both reduce the complexity of the DL model and obtain better results \cite{Martinez2021, Zhan2022}.
    
    \item \textbf{Algorithm construction}:  The pipeline is also considered part of the algorithm construction process. Areas like data preprocessing have had a great impact on this process. At a lower level, we may use an AI model to design the components of an algorithm. AutoML-zero is a notable example of this kind \cite{Real2020}. Neuroevolution  \cite{Stanley201924} is also an alternative to designing entire DL models from scratch. This has the potential to design algorithms that generalize to multiple tasks directly, but very little work is known to design algorithms from scratch to tackle new tasks.
       
\end{itemize}

The goal of enriching an AI with the input of another AI is normally to deal with intrinsic issues for models to generalize well. According to \cite{Triguero2024}, the following five enrichment approaches are key in GPAIS.

\begin{itemize}[leftmargin=*]
    \item \textbf{Discovering new behaviors} to cope with dynamic changes in the environment, such as shifts in the underlying data distribution. Continual learning \cite{Parisi2019} is a prime example of this area. 
    \item \textbf{Data generation} to mitigate the lack of data that might resemble what we may find for a new task. With the emergence of GenAI, generative models \cite{Stokel-Walker2023} may be capable of creating data that follows the distribution of a given training dataset. POET \cite{Wang2019} is a relevant example of environment/scenario generation to improve the generality of a model in potential open-world scenarios.
    \item \textbf{Learning to learn} to again compensate for the lack of data for a particular task in the presence of many (sometimes related) tasks. Learning how to transfer knowledge effectively from a set of tasks to a new task \cite{Wang2017} may enable GPAIS to generalize well. Few-shot learning \cite{Wang2020,TangXin2021} and self-learning \cite{LiuXiao2023} are notable examples of this area.
    \item \textbf{Active learning} approaches autonomously seek human assistance for additional data to perfect an existing model \cite{Fang2017}. This may provide a safer adaptability to new tasks in unfamiliar or uncertain situations.
    \item \textbf{Cooperative and collective learning} can be used to build larger systems that are composed of different AI models, enriching each other with different insights. For example, by exploiting different data modalities. 
\end{itemize}

\section{Matching the potential of EC with GPAIS: A Taxonomy for EC-powered AI and notable milestones} \label{sec:eapowerai}

As previously discussed, EC can be employed as a new layer of abstraction to facilitate the design or enhancement of AI, called EC-powered AI. This approach is of great significance for the realization of GPAIS.  The aforementioned advantages of EC regarding domain agnosticism, data assumptions, and others make it suitable for GPAIS. Figure \ref{fig:aipowerai} presents the taxonomy for this kind of approach based on \cite{Triguero2024}. To show the potential of EC with GPAIS, Subsection \ref{sec:eadesign} and \ref{sec:eaenrich} describe how EC can help to design and enrich GPAIS, respectively. Furthermore, Subsection \ref{sec:matchingEAGPAIS} analyzes the suitability of EC in GPAIS by examining its properties and their connection to research areas of ML in which EC has had a substantial contribution. Finally, Subsection \ref{sec:milestones} shows several milestones of EC-GPAIS during the last years.

\subsection{EC-powered AI for designing GPAIS} \label{sec:eadesign}

Building upon the revision established in Section \ref{sec:aipowerai} and the taxonomy presented in this section, we now explore each category and examine the strategies in which EC has made contributions, serving as a reference  for future developments of GPAIS using EC. We briefly outline and describe them to highlight the potential of EAs in design scenarios:

\begin{itemize}[leftmargin=*]
    \item \textbf{Hyper-parameter optimization:} This category has undergone extensive studies over the years, with Genetic Algorithms emerging as a widely utilized approach for addressing the problem \cite{Difrancesco2018}. However, alternative heuristics such as Particle Swarm Optimization or Bayesian Optimization also demonstrate efficacy in this context \cite{Yang2020}. In the domain of DL, Evolutionary DL typically focuses on discovering the optimal set of hyper-parameters \cite{DelSer2019,Darwish2020,Barthi2020}. Evolutionary Neural Architecture Search (NAS) methods are particularly adept for identifying the best hyper-parameter configurations for models \cite{Orive2014}. Furthermore, recent advancements in Neuroevolution have integrated NAS and co-evolutionary EAs, incorporating hyper-parameters as part of the evolutionary process \cite{Miikkulainen2019}.
\begin{figure}[t]
    \centering
    \includegraphics[width=\columnwidth]{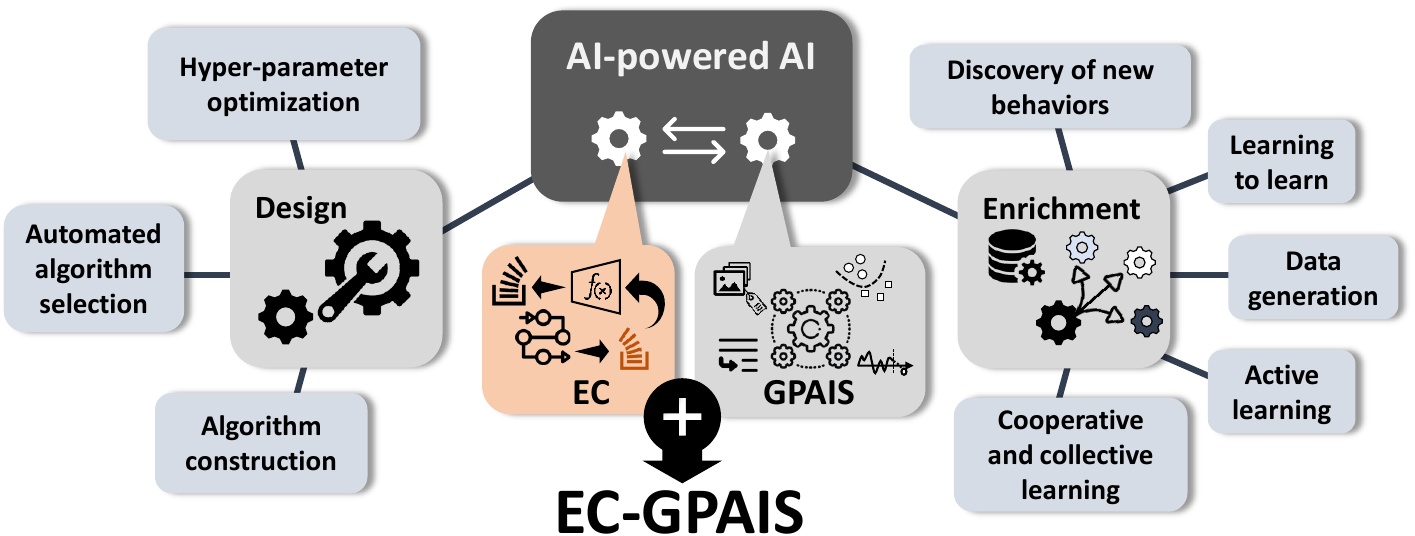}
    \caption{Taxonomy of EC-powered AI in the context of GPAIS.}
    \label{fig:aipowerai}
\end{figure}

    \item \textbf{Automated algorithm selection:} The area of Evolutionary NAS in DL has received considerable attention, with various versions of Genetic Algorithms employed to select the optimal structure for DL models \cite{XueB2020,XueB2020b}. Their capacity to explore large search spaces renders them well-suited to these tasks, as highlighted in \cite{Yuqiao2023,Zhan2022} both for single and multi-objective EAs. It is important to note that NAS often encompasses both model selection and hyper-parameter optimization, with many proposals treating them as a unified task.
    
    \item \textbf{Algorithm construction:} For an algorithm to be well constructed, all its parts should be designed consciously, from data preprocessing to the components of the ML model itself. In this context, EC has achieved great success in data preprocessing \cite{Linan2023}, including feature selection \cite{Espejo2010,XueBing2016} and feature construction \cite{Tran2016}. Another area within data preprocessing and EC is evolutionary manifold learning, which aims to construct a lower-dimensional representation (\emph{embedding}) of the structure of a dataset \cite{lensen2019can}. Techniques within this area can be regarded as a preprocessing step, as the dimensionality of data is reduced while retaining their essential features \cite{Lensen2020}. When it comes to the construction of the ML model, evolutionary DL proposals are capable of designing DL models across a wide range of domains \cite{Elsken2019b,Zhou2021,Martinez2021}. Genetic Programming has been widely used to automate the heuristic design process \cite{Burke2009,Burke2013} of most phases of a data modeling pipeline, from data preprocessing to the composition of the ML model itself \cite{Harith2019}. This family of EC solvers is at the core of AutoML-zero \cite{Real2020}, an evolutionary method capable of constructing neural networks from low-level processing primitives.   
\end{itemize}

\subsection{EC-powered AI for enriching GPAIS} \label{sec:eaenrich}

This section focuses on how EC can be exploited to enhance GPAIS. The need for generalization and adaptability capabilities of GPAIS are two key issues for which EC has demonstrated to succeed. Based on the taxonomy presented previously, enriching GPAIS using EC may be realized following these avenues: 

\begin{itemize}[leftmargin=*]
    \item \textbf{Discovering new behaviors}: For GPAIS to effectively adapt to dynamic environments, evolutionary dynamic optimization emerges as a critical area, offering innovative proposals \cite{Nguyen2012}. Additionally, when combined with multi-objective EAs, this approach further enhances adaptability \cite{Azzouz2017}.
    \item \textbf{Data generation}: In situations where GPAIS confronts limited data, the primary objective of EC is to extract high-quality insights from the available data pool. Evolutionary data generation is widely employed across various domains, including addressing imbalance problems \cite{Krawczyk2016}. Additionally, open-ended evolution not only generates data but also creates learning environments, enriching the model's understanding \cite{Wang2019}. Moreover, EC can generate diverse configurations of the initial model to adapt the model to the data \cite{Chen2023}. Finally, quality-diversity optimization offers a comprehensive approach to this category. These methods ensure the generation of a maximally diverse collection of high-performing individuals with desirable properties such as robustness and adaptability to diverse scenarios \cite{Pugh2017}.
    \item \textbf{Learning to learn}: Given that GPAIS often operates with limited data, leveraging information from similar tasks becomes crucial. Evolutionary transfer learning (ETL) and evolutionary transfer optimization (ETO) may not only pass on learned parameters but also representations and operators \cite{Tan2021}. Furthermore, the integration of evolutionary dynamic optimization with transfer learning expands the scope of applications in GPAIS scenarios \cite{Jiang2021,Wu2023}.
    \item \textbf{Active learning}: This approach has been explored in conjunction with multi-objective optimization, yielding methods capable of leveraging knowledge and discovering high-quality solutions \cite{Reyes2018}. The Pareto front delineates a region within the search space where good solutions reside, thus enriching the model by locating high-quality solutions \cite{Luo2022}. Also, integrating Active Learning with other EC-based approaches, like co-evolutionary EAs, can further enhance interactions and lead to improved solutions \cite{Lely2014}.
    \item \textbf{Cooperative and collective learning}: Evolutionary multitask optimization (EMO) \cite{Osaba2022} facilitates knowledge maximization among tasks by allowing them to learn from each other \cite{Xu2022}. EC plays a crucial role in enhancing this process, particularly in GPAIS scenarios where tasks may vary in type, enabling the EC algorithm to exploit synergistic interactions. Furthermore, co-evolutionary EAs aid in information exploitation by facilitating the exchange of information between populations \cite{Xiaoliang2019,TangXin2021}. Ensemble learning leverages collective learning of individual models to exploit the synergy between tasks \cite{heywood2023evolutionary}. This approach allows the system to harness the collective knowledge for improved performance. It is worth noting that this category not only enhances existing capabilities but also has the potential to discover new behaviors through cooperation and information exchange, further enhancing the system.
\end{itemize}

\subsection{Connection among GPAIS, Machine Learning, and Optimization Research areas} \label{sec:matchingEAGPAIS}

In this section, we analyze the capabilities of EC for GPAIS by considering their various needs. We examine which properties of GPAIS can be supported by the knowledge and achievements reported from different areas in EA research. We complement this analysis with an assessment of the competences of EC research areas for GPAIS by considering their properties:

\begin{itemize}[leftmargin=*]
\item \textbf{GPAIS need to adapt their knowledge to tasks that vary over time by exploiting previous knowledge}: For these scenarios, EC algorithms for dynamic optimization could be very useful, as it is necessary to adjust the algorithm's search strategy at runtime to adapt the search to the changing optimization landscape. In particular, they could detect new patterns over time, or decide to ignore previous patterns that no longer reflect the current tasks.

\item \textbf{GPAIS can perform several tasks simultaneously, including different priorities or objectives}: Multi-objective EAs could be used to optimize GPAIS taking into account different objectives and obtaining models with different balances among them. Also, multitask EAs can learn different tasks simultaneously, allowing for the improvement of GPAIS when tackling multiple tasks. Another area of EC that could be applied is co-evolutionary algorithms, by which many algorithms run in parallel and exchange information to improve the search. This co-evolutionary approach could help GPAIS to exploit synergies between tasks by sharing knowledge between them.

\item \textbf{GPAIS can address new unseen tasks with few or no new data}: To enforce this capability, EC can be employed for simultaneously optimizing multiple AI models, ensuring diversity among their modeled knowledge. This diversity spans a multitude of options, increasing the likelihood of identifying models that exhibit superior performance for new, unseen tasks.

\item \textbf{GPAIS should be able to construct/configure themselves autonomously}: EC has traditionally been employed for the automatic tuning of ML models across three different granularity levels. At the highest level, EC is used for algorithm selection, a process where they have seen extensive application. Moving to an intermediate level, once the algorithm is determined, EC can configure its parameters to optimize performance. Finally, at the lowest level of granularity, EC is involved in the configuration or construction of model primitives, showing their versatility across all levels of model tuning.

\item \textbf{GPAIS should run efficiently by design, both for the training phase and the inference process}: Improving the performance of EC algorithms has been widely studied with different techniques that can be transferred to other contexts. For example, to mitigate training costs, an EC algorithm can be employed to reduce the training dataset, creating a reduced dataset with similar performance — a technique known as data distillation. Additionally, EC can be used to reduce model complexity, either through pruning or quantization, thereby decreasing both training and inference costs.

\item \textbf{GPAIS should explore, evaluate, and decide on actions or sequences of actions in pursuit of specific goals or objectives}: This is a common scenario for EC. EC approaches, such as memetic computing, are a solid and robust alternative for searching in complex domains with its exploration-exploitation combination approach. Even more, EC has been traditionally used for reinforcement learning, where the outcomes of actions are used to reinforce the best actions—a technique applied across various domains from games to robotics. Consequently, EC can offer robustness and adaptability even in dynamic environments.

\item \textbf{GPAIS should simultaneously tackle multimodal modeling tasks defined over diverse datasets}: EC demonstrates versatility in handling various representations seamlessly. EC algorithms tailored for particular representations can work in cooperation to address multimodal tasks effectively. Consequently, the multi-modality of knowledge can be improved through EAs, facilitating the integration of heterogeneous information into a cohesive feature space.

\item \textbf{GPAIS should learn cooperatively, exchanging knowledge about their learned task(s) and exploiting synergies therefrom}: Numerous techniques have been developed to facilitate the exchange of information among models, such as co-evolutionary EAs or multitask optimization.

\item \textbf{GPAIS are capable of estimating their confidence when addressing their task(s) and proactively requesting new information when they are uncertain about their outcome}: In scenarios where there is a lack of sufficient information, GPAIS should autonomously decide to request additional data. However, to minimize the demand for new data and ensure its effectiveness, it should be judiciously selected and widely distributed. EAs can generate prototypes of new data and select among them based on their potential contribution to the existing dataset. Moreover, an expert can supervise and evaluate the various options proposed by the EC algorithm. This approach allows the expert to focus on evaluating the proposed options, rather than solely on proposing the need for new information.
\end{itemize} 

\begin{table*}[htb!]
    \caption{Match between GPAIS properties, ML, and EC research areas, together with the motivation for EC-GPAIS.}
    \label{tab:relationGPAISEAS}
    \centering
    \resizebox{\columnwidth}{!}{
    \begin{tabular}{L{5.3cm}L{3.2cm}L{4cm}L{7.5cm}} 
    \toprule
      \textbf{GPAIS Property}  & \textbf{ML research area} & \textbf{EC research area} & \textbf{Motivation for EC-GPAIS}\\ 
      \midrule
      GPAIS adapt their knowledge to tasks that vary over time by exploiting previous knowledge & 
      Continual learning, data shift, concept drift, transfer learning &
      Evolutionary dynamic optimization & \tabitem Appearance/disappearance of new patterns \newline\tabitem Gradual/sharp changes of tasks over time \\
      \midrule
      GPAIS can perform several tasks simultaneously, exploiting synergies among them &
      Multitask learning, transfer learning, meta-learning & 
      Multi-objective optimization, evolutionary multitask optimization, cooperative coevolution & 
      \tabitem Conflicting nature of the different objectives $\mapsto$ need for a balance between them\newline\tabitem Exchange of knowledge between GPAIS addressing different tasks \\
      \midrule
      GPAIS can address new unseen tasks in the low/null data regime &
      Zero-/Few-shot learning	&  
      Open-ended evolution, quality-diversity optimization, multimodal EAs, evolutionary data generation &
      \tabitem Modeling the unknown in open-world GPAIS $\mapsto$ diversify model's knowledge \\
      \midrule
      GPAIS should be able to construct/configure themselves autonomously & 
      Auto-ML, automated algorithm selection/design, meta-learning & 
      Evolutionary neural architecture search, genetic programming, hyper-heuristics	& 
      \tabitem Configuration in terms of low-level processing primitives\newline \tabitem Larger search spaces \\
      \midrule
      GPAIS should run efficiently at design, training \& inference & 
      ML validation methods, model compression (quantization/pruning), knowledge distillation, data distillation, structured sparsity & 
      Evolutionary quantization/pruning, large-scale global optimization, parallel EAs, distributed EAs, surrogate-assisted optimization, dynamic resource allocation in EAs &
      \tabitem Model evaluation $\mapsto$ High computational costs\newline\tabitem Task variability over time may require reconfiguration besides incremental training \\
      \midrule
      GPAIS should explore, evaluate, and decide on actions or sequences of actions in pursuit of specific goals or objectives	&
      Monte Carlo Tree Search, (multi-agent) reinforcement learning, dynamic programming, A* &
      Evolutionary reinforcement learning, evolutionary game theory, evolutionary robotics &
      \tabitem Exploration-exploitation trade-off to discover optimal strategies (exploration) while exploiting known strategies for maximizing rewards (exploitation)\newline
      \tabitem Sample efficiency over the space of possible strategies\newline
      \tabitem Robustness and adaptability in dynamic setups \\
      \midrule
      GPAIS should simultaneously tackle multimodal modeling tasks defined over diverse datasets &
      Data fusion, multimodal learning, representation learning, cross-modal retrieval, domain adaptation &
      Mixed-type EAs, evolutionary representation learning, co-evolutionary EAs, ensembles of EAs, genetic programming &
      \tabitem Integration and encoding of heterogeneous information from different modalities into a cohesive feature space\newline
      \tabitem Robust adaptability to varying data characteristics\\
      \midrule
      GPAIS should learn cooperatively, exchanging knowledge about their learned task(s) and exploiting synergies therefrom & Domain adaptation, transfer learning, federated learning & Evolutionary multitask optimization, distributed EAs, co-evolutionary EAs & \tabitem Simple knowledge exchange strategies between EAs can be applied to GPAIS modeling related tasks \\   
      \midrule
      GPAIS are capable of estimating their confidence when addressing their task(s) and proactively request new information when they are uncertain about their outcome & Uncertainty estimation, Bayesian modeling, active learning & Evolutionary data augmentation, evolutionary prototype generation, evolutionary prototype selection & \tabitem EAs can be used to optimize the selection of data points whose supervision is queried from the oracle.\newline \tabitem EAs can be used to select a diverse and representative subset of unlabeled data points for labeling.\\
     \bottomrule
    \end{tabular}}
\end{table*}

Table \ref{tab:relationGPAISEAS} summarizes the results of this analysis, matching each property of GPAIS with their related ML optimization research area, and providing a non-exhaustive listing of the EC areas of study that connect to each property. In conclusion, EC-GPAIS become the logical extension of traditional evolutionary ML, where EC has been traditionally applied to design and optimize ML systems. 

\subsection{Notable milestones and achievements in EC-GPAIS} \label{sec:milestones}

In this section, we take a closer look at specific case studies represented by proposals that have significantly contributed to the field's advancement in recent years. We illustrate different studies documented in the literature for closed- and open-world EC-GPAIS in Subsections \ref{sec:closedEAGPAIS} and \ref{sec:openEAGPAIS}, respectively. Table \ref{tab:eagpais} shows these successful proposals across different areas. The first two rows correspond to proposals for closed-world GPAIS in which there are no mechanisms for adapting to new tasks. While these proposals represent progress towards open-world GPAIS, they remain in a closed scenario. The last row, separated with a thicker line, is composed of evolutionary approaches in open-world GPAIS, because they incorporate mechanisms to generate diversity and to share knowledge. 

\subsubsection*{Current milestones in closed-world EC-GPAIS} \label{sec:closedEAGPAIS}

Proposals listed in this table have set a significant course in research, setting the grounds for initial steps in the synergy between EC and GPAIS. Many works in NAS, evolutionary DL and neuroevolution have spurred substantial developments in such areas, as evidenced by comprehensive surveys published over the years \cite{Tan2018,Harith2019,Telikani2021,Taner2022,Linan2023,Yuqiao2023,Martinez2021,Darwish2020,Zhan2022}. The common points between these proposals are the focus on the configuration of the model, by leveraging on EC to evolve weights or hyper-parameters of a neural network-based GPAIS. We begin by describing several case studies related to \textit{Hyper-parameter optimization}:
\begin{itemize}[leftmargin=*]
    \item \textbf{NEAT} \cite{Stanley2002} is the foundational stage in the field of neuroevolution. It uses an EA to evolve minimal neural networks towards the creation of deeper network architectures.
    \item \textbf{EDEN} \cite{Dufourq2017} proposes to evolve different types of convolutional, pooling, and fully-connected layers with their hyper-parameters in deep neural networks.
    \item \textbf{EvoAAA} \cite{Charte2020} is a NAS proposal linked to autoencoders. In this case, the configuration of the model (architecture, weights, and hyper-parameters) is evolved towards a network with a better accuracy performance. 
    \item \textbf{DENSER} \cite{Assunccao2019} belongs to a branch of proposals in NAS that are characterized by the use of a grammar of operators to evolve neural networks. DENSER uses a genetic approach that encodes the macro-structure of the neural network (layers, learning, parameters, etc), whereas a grammatical evolution specifies the parameters of each evolutionary algorithm unit and the valid range of the parameters.
\end{itemize}

\begin{table}[h]
    \caption{EC-based approaches for closed-world and open-world GPAIS.}
    \label{tab:eagpais}
    \centering
    \resizebox{.8\columnwidth}{!}{
    {
    \begin{tabular}{cC{2cm}C{3cm}C{3cm}C{3cm}} \toprule
      GPAIS & Objective & \makecell{\textbf{Hyper-parameter}\\\textbf{optimization}} & \makecell{\textbf{Algorithm}\\\textbf{selection}} & \makecell{\textbf{New algorithm}\\\textbf{construction}} \\ \midrule
      \multirow{2}{*}{\rotatebox[origin=c]{90}{Closed-world}} & Performance  & NEAT \cite{Stanley2002}\newline EDEN \cite{Dufourq2017}\newline EvoAAA \cite{Charte2020}\newline DENSER \cite{Assunccao2019} & LSEIC \cite{Real2017large}\newline CoDeepNEAT \cite{Miikkulainen2019}\newline LEAF \cite{Liang2019}\newline A-MFEA-RL \cite{martinez2021adaptive} &  AutoML-zero \cite{Real2020} \\ \cmidrule{2-5}
      & Performance \& Complexity & \makecell{NSGA-Net \cite{Lu2019}\\NSGANetv2 \cite{LuZhichao2020} \\NAT \cite{LuZhichao2021}} & LEMONADE \cite{Elsken2019}\newline MOENAS-TF-PSI \cite{Phan2023} & MOAZ \cite{Guha2023}\\ \midrule
      \rotatebox[origin=c]{90}{Open-world}& Diversity & POET \cite{Wang2019}\newline EGANS \cite{Chen2023}\newline EUREKA \cite{JasonMa2023}\newline XferNAS \cite{Wistuba2020}\newline ESBMAL \cite{Reyes2018} & -- & --  \\ \bottomrule
    \end{tabular}
    }}
\end{table}

The \textit{Algorithm selection} category differs from the previous category. The emphasis in this second category is placed on the search for the best algorithm rather than on the best hyper-parameter configuration. In NAS and evolutionary DL, authors often overlap these categories, as the network architecture is often encoded as part of the hyper-parameters to be optimized. Therefore, the evolution of the network architecture can yield the best \emph{algorithm} (neural network) to address the problem at hand. We highlight several works under this category that have achieved a great influence in this strand of literature:
\begin{itemize}[leftmargin=*]
    \item \textbf{CoDeepNEAT} \cite{Miikkulainen2019} represents one of the most renowned advances in the field of NAS. A co-evolutionary scheme is applied with two populations of blueprints (the backbone of the neural network) and the modules to be inserted in each part of the blueprint. Then, an evolutionary search is run toward achieving the best backbone and modules, alongside their parameters.
    \item \textbf{LSEIC} \cite{Real2017large} provides insights about the evolution of large-scale classifiers, intending to automatically search for the best architecture to address the problem at hand. To this end, this work proposes to resort to several encoding strategies and mutation operators in their evolutionary solver.
    \item \textbf{LEAF} \cite{Liang2019} is a framework based on CoDeepNeat as their inner engine to evolve networks. It constitutes an evolutionary AutoML framework that optimizes not only hyper-parameters but also network architectures and their size.
    \item \textbf{A-MFEA-RL} \cite{martinez2021adaptive} harnesses the capability of evolutionary multitask optimization to configure GPAIS capable of simultaneously solving multiple reinforcement learning environments. To this end, an evolutionary search is performed over a unified search space that represents the neural network architecture, the number of neurons of each layer, and the presence of shared layers among models.
\end{itemize}

Other NAS propose alternative metrics beyond accuracy, particularly focusing on the complexity of networks. These contributions share similar characteristics with the previous ones but mostly incorporate multi-objective evolutionary algorithms to solve for such objectives. We next summarize some of the most well-known representative approaches, which are deeply discussed on \cite{Yuqiao2023}:
\begin{itemize}[leftmargin=*]
    \item \textbf{NSGA-Net} \cite{Lu2019} represents an application of an evolutionary multi-objective solver (NSGA-II) to find neural networks that best balance between modeling performance and complexity. NSGA-Net involves an exploration of the space of potential neural network architectures in three steps: a population initialization step that is based on prior knowledge from hand-crafted architectures, an exploration step using crossover and mutation in the architectures, and an exploitation step based on a history of evaluated neural architectures.
    \item \textbf{NSGANetv2} \cite{LuZhichao2020} extends the previous NAS proposal based on a multi-objective evolutionary algorithm that uses two surrogates, one at the architecture level and another at the weights levels. In the architecture level, the surrogate improves sample efficiency. In the weights level, the weights are evolved through a Supernet based on the candidate architectures.
    \item \textbf{NAT} \cite{LuZhichao2021} presents a mechanism to automatically design neural networks by leveraging transfer learning with multiple objectives. To realize this goal, NAT utilizes task-specific Supernets that share their knowledge with other subnets within a many-objective evolutionary search process. While this approach progresses towards open-world GPAIS by featuring knowledge sharing, it embodies a closed-world GPAIS as it lacks adaptation mechanisms for new tasks. 
\end{itemize}

The next two proposals relate to \textit{Algorithm selection} with several objectives, focusing on the achievement of the network itself rather than the optimization of their weights:
\begin{itemize}[leftmargin=*]
    \item \textbf{LEMONADE} \cite{Elsken2019} uses a multi-objective evolutionary solver to search for architectures under multiple objectives. The novelty relies on how LEMONADE tackles resource consumption, using a Lamarckian inheritance mechanism. This mechanism generates warm child networks, starting with the predictive performance of their trained parents. In doing so, morphism operators are applied, using a similar concept to that used in previously explained proposals.
    \item \textbf{MOENAS-TF-PSI} \cite{Phan2023} is another multi-objective EA that aims to improve certain solutions on approximation fronts using a local search called \emph{potential solution improving}. Moreover, it resorts to a metric based on accuracy as a training-free metric to estimate the performance of the evolved network without running any training epoch, reducing the considerable computation cost that is typical of NAS methods. 
\end{itemize}

This short glimpse at the recent literature on closed-world EC-GPAIS ends by assessing efforts towards \textit{new algorithm construction}. In those cases, the goal is to create an entirely new model and its learning algorithm from low-level processing primitives. Two recent proposals fall within this category:
\begin{itemize}[leftmargin=*]
    \item \textbf{AutoML-zero} \cite{Real2020} employs an EA that disrupts the traditional paradigm in closed-world GPAIS by going beyond performance optimization. It not only discovers the optimal hyper-parameters but also develops an entire algorithm for creating a model tailored to a given modeling problem. 
    \item \textbf{MOAZ} \cite{Guha2023} represents the multi-objective variant of AutoML-zero. The goal is to find solutions over the entire Pareto front by trading off accuracy against the computational complexity of the algorithm. In addition to generating different Pareto-optimal solutions, MOAZ can effectively traverse the search space to improve search efficiency using specialized crossover and mutation operators.
\end{itemize}

These EC-GPAIS have predominantly focused on the performance-driven optimization of GPAIS and on the addition of additional search objectives, such as complexity. However, there is still a road ahead to make these EC-GPAIS compliant with their assumed properties, especially in what refers to their multimodal, multitask nature and their computational efficiency. Closed-world achievements reported to date only consider one single task, require high computational costs, and assume enough available data for the formulation of an objective function that prevents the optimized GPAIS from overfitting.

\subsubsection*{Current milestones in open-world EC-GPAIS} \label{sec:openEAGPAIS}

More recent contributions have shifted their scope towards guaranteeing that GPAIS can integrate new task(s) into their learned knowledge, different from the traditional objective-driven search methods that limit the ability of a system to integrate a new task. While there may not be specific surveys on these emerging diversity-driven concepts, several research areas are closely aligned with the principles of open-world GPAIS. The last studies revisited in what follows are precisely based on the generation of diversity and the discovery of diverse model behaviors:
\begin{itemize}[leftmargin=*]
    \item \textbf{POET} \cite{Wang2019} focuses on generating diversity via data synthesis. In this work, agents are paired with these newly generated environments to learn from them and, at the same time, the agents' weights are also optimized. Moreover, these agents can even be transferred from one environment to another, using their knowledge to adapt to the other. Diversity is induced through data synthesis, where knowledge is transferred between the environments via the agents, ultimately optimizing the model through the agents' weights. 
    \item \textbf{EGANs} \cite{Chen2023}, framed in the area of Generative Adversarial Networks, serves as an example of generating diversity through the model, particularly in a zero-shot learning environment. By using an evolutionary approach, EGANs initially learn the optimal generator of models. In a subsequent stage, this generator becomes part of another evolutionary process to determine the final model. 
    \item \textbf{EUREKA} \cite{JasonMa2023} constitutes another open-world GPAIS in which several reinforcement learning tasks are performed at the same time. The evolutionary algorithm evolves several reward functions in a context based on the environment source code. EUREKA generates executable reward functions, improving them with an evolutionary search that iteratively produces batches of reward candidates. 
    \item \textbf{XferNAS} \cite{Wistuba2020} introduces an open-world GPAIS framework for knowledge transfer. XferNAS collects source knowledge from multiple tasks and combines this knowledge to generate an architecture for a new task. 
    \item \textbf{ESBMAL} \cite{Reyes2018} integrates Active Learning and EAs for improving data labeling. The EA optimally reports batches of data to enhance the instance selection process, contributing to the set of open-world proposals together with the aforementioned approaches.
\end{itemize}

In our research, we have not encountered any specific work focused on the utilization of EC for algorithm selection or construction in open-world GPAIS. Differently from closed-world GPAIS, the use of new diversity metrics to develop open-world GPAIS can be regarded as a research niche yet to be explored. 

Another important feature is that current GPAIS usually consider one task in a closed-world setting, which is even rarer in open-world problems. Surprisingly, the adoption of EC has been used for the design and optimization of multitask learning models \cite{Zhao2023}. We have also highlighted EUREKA as an open-world GPAIS proposal that works with reinforcement learning agents capable of doing several tasks. These detected research niches should stimulate efforts in prospective studies related to GPAIS and EC.

GenAI systems are usually more autonomous and do not depend on experts. Still, the use of EC can allow them to achieve even higher levels of autonomy, further reducing the need for such an expert. Using another AI to help improve it can be a viable approach. Within the field of GenAI, more and more works are using EC to improve such systems, which is a major focus in the coming years \cite{Wu2024}. In this research line, recent works leverage EAs as key algorithms for tasks such as model merging and creation of foundation models \cite{Akiba2024}, evolving code generated by Large Language Models \cite{Hemberg2024}, evolving prompts of Large Language Models \cite{Guo2024}, and generating optimization algorithms \cite{Liu2023b}.

\section{Challenges of harnessing the EC-GPAIS benefits and strategies to address advances} \label{sec:challenges}

\begin{figure*}
    \centering
    \includegraphics[width=\columnwidth]{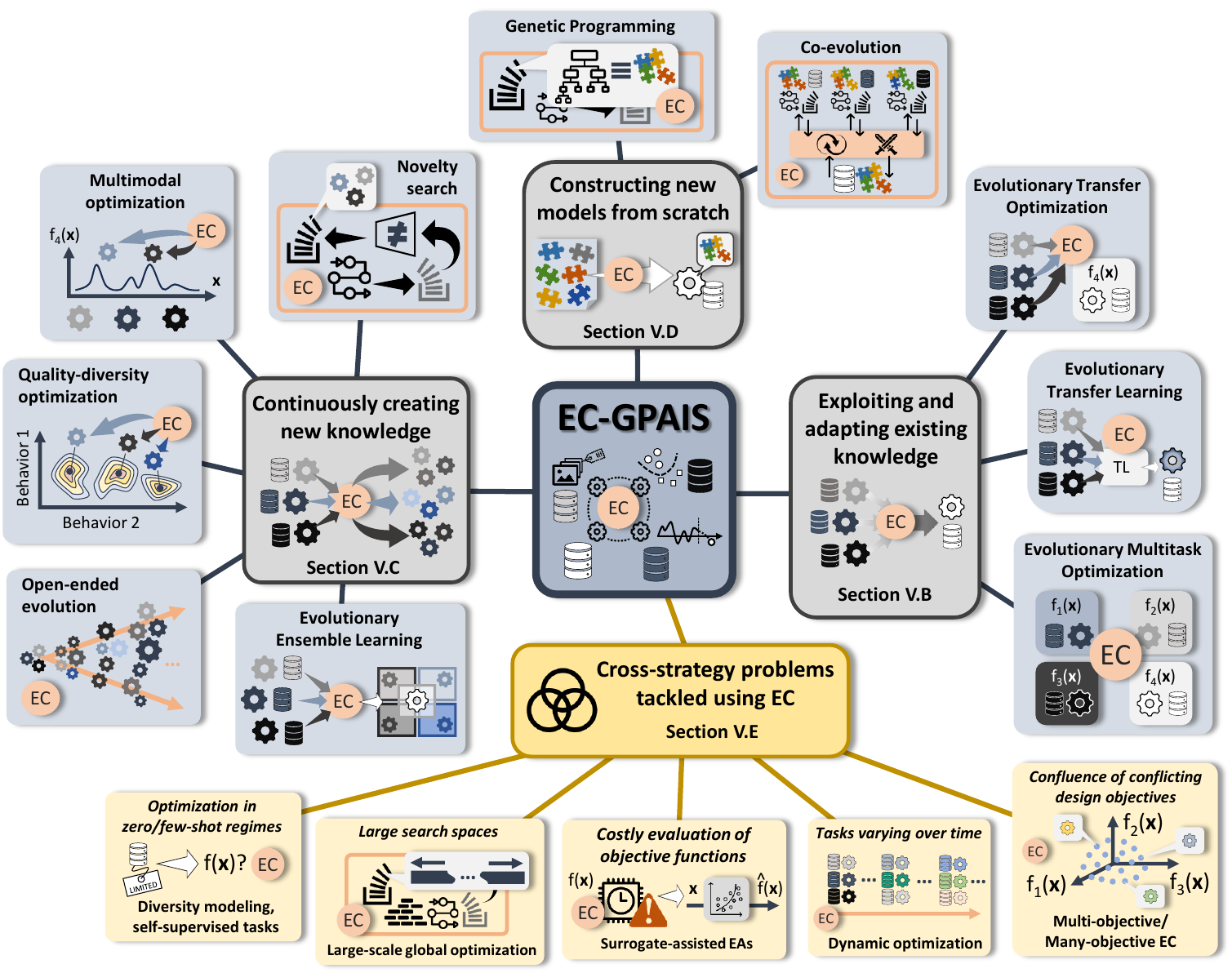}
    \caption{Summary of the different strategies to leverage EC in the design and enrichment of GPAIS, together with the EC research areas related to each strategy. Problems undergone by the adoption of EC are common to all strategies, and the research areas studying how to overcome them are also included.}
    \label{fig:strategies}
\end{figure*}

The integration of EC with GPAIS poses significant challenges, despite their potential benefits. A major difficulty is ensuring synergy between the adaptive nature of EC and the complex decision-making processes of GPAIS. For this reason, we recall these challenges of how EC is capable of adapting to GPAIS for the design and optimization of these systems in Subsection \ref{sec:motea}. To overcome these challenges, we also present several strategies that can be implemented with EC to both design and improve GPAIS. Specifically, we focus on their goal, their importance in the context of GPAIS, and an analysis of the EC-based research areas that can help realize these strategies. This analysis, backed by the graphical summary in Figure \ref{fig:strategies} and through Sections \ref{sec:eastrategy_a} to \ref{sec:eastrategy_d}, serves as a motivating evidence for the present and future development of EC-GPAIS.

\subsection{Challenges of harnessing the benefits of EC-GPAIS} \label{sec:motea}
 
There are many ways in which EC can benefit both closed-world and open-world GPAIS. EC can enhance closed-world GPAIS by facilitating advanced data preprocessing, optimizing hyper-parameters, and adapting models. Additionally, EMO uses EC algorithms to simultaneously learn multiple tasks, showcasing their versatility in addressing various challenges individually.
 
In open-world scenarios, models must adapt to new problems with minimal data, ensuring robust performance across various challenges. EC offers flexible fitness functions, enforcing robustness within the solution space. Multifactorial optimization supports robustness across tasks, potentially improving performance on new problems. 

EC offers the advantage of generating multiple solutions throughout the optimization process, ensuring a range of optimized AI models to tackle new problems. Using a variety of AI models can improve results, as different models may perform better for specific challenges. Maintaining this diversity is important for achieving competitive results without extensive retraining.

\subsection{EC-GPAIS by exploiting and adapting existing knowledge} \label{sec:eastrategy_a}

\subsubsection*{Why is this strategy important for GPAIS?} The development of a system capable of performing a diverse set of tasks requires a significant effort. Therefore, every piece of knowledge within the system should be regarded as crucial for its continuous improvement. GPAIS need to leverage all their knowledge from the existing pool of tasks to enhance performance and adapt to new incoming tasks. This forces the system to maintain their performance level on previous tasks, while successfully adapting to new ones appearing over time.

\subsubsection*{How can EC support this strategy?} As stated before, ETO \cite{Tan2021}, EMO \cite{Osaba2022}, and ETL \cite{Zhuang2021} have shown to be strategies to optimize previously acquired knowledge. Therefore, by generalizing learning across problems, ETO, EMO, and ETL can be useful to optimize the exchange of knowledge among models devised for different tasks. 

\subsubsection*{Which EC research areas are useful for this strategy?} The aforementioned research areas in EC can be linked to the strategy of exploiting and adapting existing knowledge: 
\begin{itemize}[leftmargin=*]
    \item ETO combines principles from EAs and transfer learning at its core, knowledge or solutions learned from an optimization problem in a source domain to improve an optimization process in another related domain. When solutions to such problems represent the knowledge of the GPAIS (i.e. its learned parameters), ETO can optimize the adaptation of the transferred knowledge to the target problem by fine-tuning or re-configuring it to better suit the characteristics of the target domain.
    \item ETL aims to leverage knowledge from related domains to improve performance in a different domain, but focuses on improving the learning process of the model using EC through transfer learning based on models, features and/or instances. ETL has been widely used in NAS, with works like XferNAS, in which a process of transfer learning is applied to initiate architectures for a new task, so the knowledge in other related problems is transferred to create a warm-started network \cite{Wistuba2020}. Another work elucidating the potential of ETL is presented in \cite{Poyatos2023}, where the process of knowledge transfer between tasks in similar domains is expedited by pruning the unnecessary components of the neural network using EAs, so the GPAIS can autonomously generalize to other tasks while retaining knowledge from the learned source tasks.
    \item Differently, EMO can be used in scenarios where the parameters of the GPAIS are evolved jointly within a single evolutionary search based on several objective functions, as has been done in \cite{martinez2021adaptive} in the context of multitask reinforcement learning. In this case, a unified search space can effectively encode parameters shared by all tasks. Evolutionary operators suited to exchange genotypic information across optimization tasks (such as those defined in multifactorial optimization \cite{Bali2020}) effectively implement knowledge transfer across tasks.
\end{itemize}  

In these three research areas, the transfer of parameters, backbones, or other architectural elements may expedite the learning process and boost even further the performance of the model. When dealing with GPAIS in these areas, the large search space of decision variables evolved by the EA can be a challenge. Recently, multitask learning approaches have been using multi-objective EAs and large-scale global optimization to ensure scalability for realistic model sizes \cite{Liu2023}. The assessment of knowledge transfer between tasks often requires training on the target task(s), resulting in significant delays. To reduce this cost, surrogates have been used in evolutionary DL \cite{XueB2020c}. In multitask GPAIS, models for different tasks may have different processing latencies, and surrogates may benefit from those learned for the fastest evaluated tasks \cite{Wang2022}.

\subsection{EC-GPAIS by continuously creating new knowledge} \label{sec:eastrategy_b}

\subsubsection*{Why is this strategy important for GPAIS?} GPAIS need adaptability to new tasks with minimal data, therefore requiring the continuous creation of diverse knowledge to accelerate responsiveness and facilitate task integration. When GPAIS have good performance over a given set of tasks (closed-world scenario), their transition to an open-world setting is delivered by using mechanisms of knowledge diversification, which can be realized through ensemble modeling, the randomization of the GPAIS parameters or the generation of synthetic data, among other strategies.

\subsubsection*{How can EC support this strategy?} EC can be leveraged for the generation of diverse models, for handling problems in different search spaces, and for the generation of synthetic data \cite{correia2023evolutionary}. The conundrum when implementing this strategy with EC is twofold: 1) how to formulate objective functions that properly model the diversity in open-world settings, even in the absence of data from the new task(s); 2) how to retain solutions during the evolutionary search that are both good for the source tasks and diverse to potentially be useful to address new tasks.

\subsubsection*{Which EC research areas are useful for this strategy?} Different EC research areas have paid attention to the discovery and retention of diverse solutions during the search. Furthermore, EC can also be useful to induce this diversity by synthesizing data that produce diverse model behaviors through training. We will now revisit some of these areas:
\begin{itemize}[leftmargin=*]
    \item \textit{Open-ended evolution} aligns with this strategy closely, as it pursues the continuous generation of knowledge without reaching an optimal state. The integration of open-ended evolution with EC makes it feasible to create systems that can continuously adapt \cite{Taylor2019}, where the EC algorithm will help to create more data or working environments \cite{Lehman2011}. This is the approach followed in POET \cite{Wang2019} or Minimal Criterion Coevolution \cite{Brant2017}.
    \item \textit{Quality-diversity optimization} aims at developing EC algorithms that prioritize generating diverse, high-quality solutions. In addition to evaluating solution quality, diversity metrics measure how distinct each solution is from others. Promoting diversity ensures that solutions are both good and varied across possible behaviors. The EC designed under this paradigm is well-suited for evolving GPAIS in open-world settings. By evolving their parameters and/or configuration and devising measures to quantify the diversity of the evolved GPAIS, quality-diversity optimization can provide a pool of diverse, well-performing models. This can offer better guarantees of adaptability to new tasks, even in the presence of conflicting objectives \cite{Neumann2019}.
    \item \textit{Multimodal optimization} also seeks to retain diverse, high-quality solutions during the evolutionary search \cite{preuss2015multimodal}. However, differently from quality-diversity optimization, the diversity is defined over the genotype of the solutions evolved by the algorithm rather than on a behavioral space. In any case, multimodal optimization can also be a promising EC research area to produce diverse models, particularly over combinatorial search spaces representing model configurations \cite{Tanabe2020}.    
    \item \textit{Novelty Search} involves driving the search not based on an objective function, but rather by a measure of the novelty of solutions in the search space \cite{lehman2011novelty,Lehman2011}. This ensures the discovery of new solutions and the accumulation of more knowledge about the problem at hand. Although both novelty search and quality-diversity optimization strive to retain diverse solutions, novelty search centers on promoting novelty as a measure of diversity, while quality-diversity optimization aims for a balance between high-quality solutions and diversity, encompassing multiple aspects beyond novelty alone. These techniques can complement each other towards creating new knowledge for open-world GPAIS, and aspects of novelty search can contribute to diversity within the framework of quality-diversity optimization. Novelty search with EC has been explored in robotics and autonomous driving systems to ensure that the created knowledge avoids system malfunctions, showcasing the possibilities brought by this EC research area for complex systems to accommodate unknown circumstances \cite{Langford2019}.
    \item \textit{Evolutionary ensemble learning} studies the optimization of the learners within an ML ensemble based on different objectives. Most literature focuses on the ensemble's performance in modeling tasks, but some works address different objectives like fairness \cite{ZhangQingquan2023}, explainability \cite{WangXin2024} and dealing with conflicting rewards in multi-agent reinforcement learning systems \cite{bai2023evolutionary}. We envision that the rich background on methodologies to model diversity in evolutionary ensemble learning (recently revisited in \cite{heywood2023evolutionary}) can open up complementary ways to realize open-world GPAIS based on ML ensembles.  
\end{itemize}

\subsection{EC-GPAIS by constructing new models from scratch} \label{sec:eastrategy_c}

\subsubsection*{Why is this strategy important for GPAIS?} Building new models from scratch is another strategy used in GPAIS. In such cases, the construction of these models is typically associated with closed-world GPAIS, as it requires the assumption of having quality data to formulate an objective function that guides the search for the optimal model for the task(s) at hand. When new task(s) arrive, the model may not be optimal any longer, so it must be optimized again for such new task(s). To address this as an open-world problem, it is crucial to modify each part of the model for the new task. However, the lack of sufficient high-quality data makes this adaptation challenging. There might not be enough prior knowledge of the new tasks to accurately assess the generalization of the developed algorithm. Consequently, the difficulty lies in adapting this approach to open-world scenarios.

\subsubsection*{How can EC support this strategy?} EC can optimize various stages of an ML pipeline, making them suitable for GPAIS. ML areas like AutoML, hyper-parameter optimization, and feature engineering have been effectively addressed via EAs \cite{Telikani2021}. GPAIS can inherit this successful history of achievements, facing new challenges such as the increased granularity at which GPAIS are constructed (e.g., low-level processing primitives, as in AutoML-zero \cite{Real2020}), or the aforementioned difficulty of formulating an optimization objective over scarce data or even no knowledge about the new task(s). 

\subsubsection*{Which EC research areas are useful for this strategy?} Areas such as AutoML and \textit{Algorithm Selection} have already been incorporated in GPAIS using EAs \cite{Liang2019}. Furthermore, AutoML can identify optimal pipelines for DL approaches, which cover feature engineering, hyper-parameter optimization, and NAS. Consequently, EC has played an important role in constructing models from scratch in recent years. This commitment is evident in various surveys on NAS that explore the construction of models \cite{Yuqiao2023,DelSer2019,Darwish2020,Zhan2022}, and also several closed-world GPAIS that fall into this strategy: AutoML-zero and MOAZ, its extension to multi-objective optimization. Lastly, in the neuroevolution field, CoDeepNEAT \cite{Liang2019,Miikkulainen2019} can be regarded as a construction of the model, despite being classified in the \textit{algorithm selection} category. This approach aims to identify the primitives for building a neural network. It involves selecting and evolving two populations: one for the network backbones and the other for their primitives.

\subsection{Cross-strategy problems tackled using EC} \label{sec:eastrategy_d}

We now delve into the usage of EC to confront and resolve cross-strategy problems within the GPAIS context. In particular, we explore how EC can effectively navigate complex problem landscapes by amalgamating various strategies to adapt to diverse conditions and optimize solutions collaboratively:

\subsubsection*{Scarcity of quality data for target tasks}

In the context of GPAIS, the primary objective is to facilitate system adaptation upon the appearance of a new task. However, the available data for this adaptation may often be insufficient. To address this, various strategies can be used synergistically, such as data synthesis and active learning. Combining these strategies with EC can reduce the time and improve the quality of adaptation. EAs, as in the case of POET \cite{Wang2019}, not only can generate similar data but also optimize batches of this new data for the system to perform better across known tasks, or adapt to new tasks in a faster and more effective fashion.

\subsubsection*{Formulation of objective functions in the zero-shot open-world regime}

In EC-GPAIS, the careful selection of objective functions is crucial for the design and enrichment of the system. As stated previously, incomplete knowledge of the target tasks may prevail during the training phase of the GPAIS, so it becomes necessary to prepare GPAIS for new tasks. ML paradigms like few-shot learning, meta-learning, or transfer learning, which can adapt the model to new tasks with limited samples of information, are often employed in zero-shot open-world scenarios. However, when tackling them with EC-GPAIS, it becomes crucial to formulate optimization goals capable of measuring the adaptability of the optimized GPAIS to the unknown. Diversity measures, the discovery of auxiliary self-supervised tasks that the GPAIS can learn preemptively, or strategies from EC-based open-ended evolution can be interesting research directions for this purpose.

\subsubsection*{Dimensionality of the search space}

The design of a system involves critical decisions about its parameters. When dealing with an extremely large system, making these choices becomes particularly challenging. Consequently, the selection of all relevant components during the design of GPAIS poses a formidable task. Within this context, large-scale global optimization emerges as a field well-suited for tackling problems characterized by high-dimensional spaces. EC has found widespread applications in large-scale optimization problems \cite{omidvar2021review}. By using EC algorithms for high-dimensional problems and knowledge transfer for open-world scenarios, we can make the design or even the construction of large GPAIS feasible and computationally affordable.

\subsubsection*{Costly evaluation of objective functions}

The selection of a model defining a given system is a pivotal decision that influences the system's performance when addressing a certain task. When this selection is performed by an EC algorithm, the evolutionary search requires evaluating the quality of multiple candidate models which, in the context of ML tasks, imply repeated training/validation passes. While the design of the model, especially for larger models, may pose a bottleneck in the evaluation process, it is crucial to note that in ML, the evaluation of the quality of such models also contributes significantly to the evaluation time. Objective functions with high computational costs can gain the computational benefits sought by adopting EC for the design and enrichment of GPAIS. Surrogate-assisted optimization is a solution to expedite evaluation times by using surrogate models to make the process more efficient and explore more candidate models. This, in turn, enhances the overall efficiency of EC-GPAIS. In tackling the interplay between model choice and evaluation efficiency, EC and surrogate modeling play a pivotal role. EC contributes to global exploration within the search space, while surrogate models enhance the computational efficiency of the overall process, streamlining the optimization process. This collaborative approach accelerates convergence toward optimal solutions, making it a compelling strategy for refining and optimizing GPAIS.

\subsubsection*{Confluence of conflicting design objectives}

When designing or improving GPAIS, a primary objective is to optimize their performance across different tasks and environments. As shown in Table \ref{tab:relationGPAISEAS}, GPAIS should not only perform well over their known tasks, but also run efficiently in terms of computational requirements, and adapt reliably to new modeling problems. Therefore, such goals posed by the design and improvement of GPAIS require balancing various design factors, such as modeling performance and computational efficiency. These design factors become in conflict with each other when decision variables refer to the architectural and/or trainable parameters of the model, as exemplified by those GPAIS approaches listed in Table \ref{tab:eagpais} under the \emph{Performance \& Complexity} objective category. Likewise, other design objectives can also be formulated and considered in the design and optimization of GPAIS: among them, explainability \cite{carmichael2021learning} and robustness to adversarial attacks \cite{liu2021multi,cheng2023neural} or out-of-distribution inputs to the model \cite{bai2021ood,gambella2024flatnas} have been explored as additional objectives for optimizing its configuration. The current global concern about the safety and trustworthiness of modern AI systems (including GPAIS) \cite{bengio2024managing} suggests that such objectives will gain increasing relevance in future studies. Given these prospects, we envision that multi- and many-objective EC will be crucial for the design and improvement of GPAIS due to its ability to optimize multiple conflicting objectives simultaneously. We have so far observed a growing trend towards considering several design goals in both closed-world and open-world GPAIS, which we expect to permeate into all strategies to realize EC-GPAIS. Possibilities include the creation of synthetic data by balancing between diversity and fidelity, EC approaches to multitask learning that account for several task-dependent goals, or multi-objective approaches to open-ended evolution. We foresee that multi-/many-objective EC techniques capable of efficiently discovering GPAIS that best balance competing objectives will become increasingly adopted in future studies.

\vspace{3mm}
In conclusion, the research areas depicted in this section delineate a promising landscape for future studies in GPAIS development. These identified EC research areas, in conjunction with those already actively contributing to the ongoing advancement of GPAIS, hold the potential to foster the emergence of systems capable of self-constructing and/or self-adapting themselves to new task(s), thus bringing AI closer to a more general AI.

\section{Conclusions} \label{sec:conclusions}

Evolutionary computation plays a significant role in the design and enhancement of AI systems. The integration of EC with GPAIS is currently leading to significant progress, as evidenced in this study. The primary objectives of this work have been: (1) introduce the term EC-GPAIS and a taxonomy for EC-powered AI, (2) analyze the diverse options for designing and enhancing GPAIS using the taxonomy of EC-powered AI, match key properties of GPAIS to EC research areas that can address them, and (3) discuss the challenges of harnessing the EC-GPAIS benefits and strategies to address advances EC-GPAIS. EC contributes to GPAIS design by optimizing hyper-parameters and configuration, selecting appropriate algorithms, or even crafting radically new algorithms. The enhancement of GPAIS through EC involves inducing diversity via evolutionary mechanisms within the GPAIS, evolving data from which GPAIS learns, or optimizing knowledge transfer across tasks.

Numerous proposals regarding closed-world GPAIS in DL exist, but it is important to acknowledge that other domains have also significantly impacted the evolution of such systems. The gap towards open-world GPAIS is narrowing, demonstrated by proposals illustrating the utilization of diversity to generate knowledge and adapt to the unknown. This notion is inspiring researchers to explore new horizons, combining their expertise with EAs to develop open-world GPAIS. With this growing interest, we anticipate a future brimming with versatile EA-GPAIS models that will serve as benchmarks in their respective fields.  This is reflected in novel visions that connect GPAIS such as Large Language Models and GenAI with emerging Artificial General Intelligence \cite{Qu2024integration,Zhang2023one}.

From the perspective of the governance of GPAIS, the inclusion of EC in the design and enrichment of these complex systems can provide further advantages, as they provide means to consider optimization objectives that measure aspects related to the trustworthiness of these systems. A general discussion about these topics has been held in \cite{Triguero2024}.

On a closing note, the vibrant research activity on GPAIS and evolutionary GPAIS noted in the last few years exposes the great expectation posed on this field to revolutionize the field of evolutionary ML, leading to more general AI systems capable of solving a wide range of tasks, and self-adapting their knowledge to tackle new ones. The flexibility and ease of adaptation of EC make them a perfect match to cope with the stringent properties sought for GPAIS, including the multimodality of the tasks being solved, their variability over time, and the large dimensionality of the design and construction of GPAIS. The hybridization of this family of solvers with GPAIS promises a bright future in the field of AI, representing the cutting edge of research in the coming years, with the ultimate goal of developing human-like systems capable of analyzing, learning, and performing several tasks on their own.

\section*{Acknowledgments}

F. Herrera, D. Molina, I. Triguero, S. Garcia acknowledge funding support by Grant ``Convenio de Colaboración entre la Universidad de Granada y la S.M.E Instituto Nacional de Ciberseguridad de España M.P., S.A. para la Promoción de Proyectos Estratégicos de Ciberseguridad en España", funded by S.M.E. Instituto Nacional de Ciberseguridad de España M.P. S.A. (``INCIBE") and European Union – NextGenerationEU.

\bibliographystyle{unsrt}  
\bibliography{references}

\begin{thebibliography}{100}

\bibitem{Song2019}
Heda Song, Isaac Triguero, and Ender {\"O}zcan.
\newblock A review on the self and dual interactions between machine learning
  and optimisation.
\newblock {\em Prog. in Artif. Intell.}, 8(2):143--165, April 2019.

\bibitem{Yang2020}
Li~Yang and Abdallah Shami.
\newblock On hyperparameter optimization of machine learning algorithms:
  {T}heory and practice.
\newblock {\em Neurocomputing}, 415:295--316, November 2020.

\bibitem{Yao2023robust}
Danial Yazdani et~al.
\newblock Robust {O}ptimization {O}ver {T}ime: {A} {C}ritical {R}eview.
\newblock {\em IEEE Trans. on Evol. Comput.}, 2023.
\newblock {I}n Press, 10.1109/TEVC.2023.3306017.

\bibitem{Goodfellow2016}
Ian Goodfellow, Yoshua Bengio, and Aaron Courville.
\newblock {\em Deep learning}.
\newblock MIT press, 2016.

\bibitem{Mikolov2013}
Tomas Mikolov et~al.
\newblock Distributed {R}epresentations of {W}ords and {P}hrases and {T}heir
  {C}ompositionality.
\newblock In {\em Advances in Neural Information Processing Systems},
  volume~26, page 3111–3119, December 2013.

\bibitem{Hirschberg2015}
Julia Hirschberg and Christopher~D. Manning.
\newblock Advances in natural language processing.
\newblock {\em Sci.}, 349(6245):261--266, July 2015.

\bibitem{Vandis2023}
E.A.M. {van Dis} et~al.
\newblock Chat{GPT}: five priorities for research.
\newblock {\em Nat.}, 614(7947):224--226, February 2023.

\bibitem{Lehman2024}
Joel Lehman et~al.
\newblock Evolution {T}hrough {L}arge {M}odels.
\newblock In {\em Handbook of Evolutionary Machine Learning}, pages 331--366.
  Springer Nature Singapore, Singapore, 2024.

\bibitem{Triguero2024}
Isaac Triguero et~al.
\newblock General {P}urpose {A}rtificial {I}ntelligence {S}ystems {{(GPAIS)}}:
  {P}roperties, definition, taxonomy, societal implications and responsible
  governance.
\newblock {\em Inf. Fusion}, 103:102135, March 2024.

\bibitem{anderljung2023frontier}
Markus Anderljung et~al.
\newblock Frontier {{AI}} regulation: Managing emerging risks to public safety.
\newblock arXiv:2307.03718, 2023.

\bibitem{bengio2024managing}
Yoshua Bengio et~al.
\newblock Managing extreme {AI} risks amid rapid progress.
\newblock {\em Science}, page eadn0117, May 2024.

\bibitem{Sun2020}
Shiliang Sun, Zehui Cao, Han Zhu, and Jing Zhao.
\newblock A {S}urvey of {O}ptimization {M}ethods {F}rom a {M}achine {L}earning
  {P}erspective.
\newblock {\em IEEE Trans. on Cybern.}, 50(8):3668--3681, August 2020.

\bibitem{Gutierrez2023}
Carlos~I Gutierrez et~al.
\newblock A {P}roposal for a {D}efinition of {G}eneral {P}urpose {A}rtificial
  {I}ntelligence {S}ystems.
\newblock {\em Digit. Soc.}, 2(36):1--8, September 2023.

\bibitem{Uuk2023}
Risto Uuk, Carlos~Ignacio Gutierrez, and Alex Tamkin.
\newblock Operationalising the {D}efinition of {G}eneral {P}urpose {{AI}}
  {S}ystems: {A}ssessing {F}our {A}pproaches.
\newblock {\em Available at SSRN}, 4471151:1--14, June 2023.

\bibitem{Campos2023}
Simeon Campos and Romain Laurent.
\newblock A {D}efinition of {G}eneral-{P}urpose {AI} {S}ystems: {M}itigating
  {R}isks from the {M}ost {G}enerally {C}apable {M}odels.
\newblock {\em Available at SSRN}, 4423706:1--8, April 2023.

\bibitem{Barrett2023}
ANTHONY~M Barrett et~al.
\newblock {{AI}} risk-management standards profile for general-purpose {{AI}}
  systems {{(GPAIS)}} and foundation models.
\newblock Technical report, Center for Long-Term Cybersecurity, November 2023.

\bibitem{Standish2003}
Russell~K Standish.
\newblock Open-ended artificial evolution.
\newblock {\em Int. J. of Comput. Intell. and Appl.}, 3(02):167--175, June
  2003.

\bibitem{Taylor2016}
Tim Taylor et~al.
\newblock {O}pen-{E}nded {E}volution: {P}erspectives from the {{OEE}}
  {W}orkshop in {Y}ork.
\newblock {\em Artif. Life}, 22(3):408--423, August 2016.

\bibitem{Packard2019}
Norman Packard et~al.
\newblock An {O}verview of {O}pen-{E}nded {E}volution: {E}ditorial
  {I}ntroduction to the {O}pen-{E}nded {E}volution {{II}} {S}pecial {I}ssue.
\newblock {\em Artif. Life}, 25(2):93--103, May 2019.

\bibitem{Zhang2023omni}
Jenny Zhang, Joel Lehman, Kenneth Stanley, and Jeff Clune.
\newblock {{OMNI}}: {O}pen-endedness via {M}odels of human {N}otions of
  {I}nterestingness.
\newblock arXiv:2306.01711, 2023.

\bibitem{Pugh2017}
Justin~K. Pugh, Lisa~B. Soros, and Kenneth~O. Stanley.
\newblock Quality {D}iversity: {A} {N}ew {F}rontier for {E}volutionary
  {C}omputation.
\newblock {\em Front. in Robot. and AI}, 3:1--40, July 2016.

\bibitem{Cully2018}
Antoine Cully and Yiannis Demiris.
\newblock Quality and {D}iversity {O}ptimization: {A} {U}nifying {M}odular
  {F}ramework.
\newblock {\em IEEE Trans. on Evol. Comput.}, 22(2):245--259, April 2018.

\bibitem{Bradley2023qd}
Herbie Bradley et~al.
\newblock Quality-{D}iversity through {{AI}} {F}eedback.
\newblock In {\em Second Agent Learning in Open-Endedness Workshop}, pages
  1--112, October 2023.

\bibitem{Lehman2008}
Joel Lehman and Kenneth~O Stanley.
\newblock Exploiting open-endedness to solve problems through the search for
  novelty.
\newblock In {\em Artificial Life XI: Proceedings of the 11th International
  Conference on the Simulation and Synthesis of Living Systems}, pages
  329--336, August 2008.

\bibitem{Harith2019}
Harith Al-Sahaf et~al.
\newblock A survey on evolutionary machine learning.
\newblock {\em J. of the R. Soc. of N. Z.}, 49(2):205--228, May 2019.

\bibitem{Telikani2021}
Akbar Telikani, Amirhessam Tahmassebi, Wolfgang Banzhaf, and Amir~H. Gandomi.
\newblock Evolutionary {M}achine {L}earning: {A} {S}urvey.
\newblock {\em ACM Comput. Surv.}, 54(8):1--35, October 2021.

\bibitem{Taner2022}
Hamit~Taner {\"U}nal and Fatih Ba{\c{s}}{\c{c}}ift{\c{c}}i.
\newblock Evolutionary design of neural network architectures: a review of
  three decades of research.
\newblock {\em Artif. Intell. Revi.}, 55:1723–1802, July 2022.

\bibitem{Linan2023}
Nan Li et~al.
\newblock Automatic design of machine learning via evolutionary computation: A
  survey.
\newblock {\em Appl. Soft Comput.}, 143:110412, August 2023.

\bibitem{Yao1999}
Xin Yao.
\newblock Evolving artificial neural networks.
\newblock {\em Proc. of the IEEE}, 87(9):1423--1447, September 1999.

\bibitem{Akiba2024}
Takuya Akiba et~al.
\newblock Evolutionary {O}ptimization of {M}odel {M}erging {R}ecipes.
\newblock arXiv:2403.13187, 2024.

\bibitem{XueBing2016}
Bing Xue, Mengjie Zhang, Will~N. Browne, and Xin Yao.
\newblock A {S}urvey on {E}volutionary {C}omputation {A}pproaches to {F}eature
  {S}election.
\newblock {\em IEEE Trans. on Evol. Comput.}, 20(4):606--626, August 2016.

\bibitem{Mauceri2021}
Stefano Mauceri, James Sweeney, Miguel Nicolau, and James McDermott.
\newblock Feature extraction by grammatical evolution for one-class time series
  classification.
\newblock {\em Genet. Program. and Evol. Mach.}, 22(3):267--295, April 2021.

\bibitem{heywood2023evolutionary}
Malcolm~I. Heywood.
\newblock Evolutionary {E}nsemble {L}earning.
\newblock In {\em Handbook of Evolutionary Machine Learning}, pages 205--243.
  Springer Nature Singapore, Singapore, 2024.

\bibitem{Fu2019}
Sibao Fu, Yongwu Li, Shaolong Sun, and Hongtao Li.
\newblock Evolutionary support vector machine for {{RMB}} exchange rate
  forecasting.
\newblock {\em Phys. A: Stat. Mech. and its Appl.}, 521:692--704, May 2019.

\bibitem{Barros2012}
Rodrigo~Coelho Barros, Márcio~Porto Basgalupp, André C. P. L.~F. de~Carvalho,
  and Alex~A. Freitas.
\newblock A {S}urvey of {E}volutionary {A}lgorithms for {D}ecision-{T}ree
  {I}nduction.
\newblock {\em IEEE Trans. on Syst., Man, and Cybern., Part C (Appl. and
  Rev.)}, 42(3):291--312, May 2012.

\bibitem{DelSer2019}
Javier {Del Ser} et~al.
\newblock Bio-inspired computation: Where we stand and what's next.
\newblock {\em Swarm and Evol. Comput.}, 48:220--250, August 2019.

\bibitem{Zhan2022}
Zhi-Hui Zhan, Jian-Yu Li, and Jun Zhang.
\newblock Evolutionary deep learning: {A} survey.
\newblock {\em Neurocomputing}, 483:42--58, April 2022.

\bibitem{Gen2023}
Mitsuo Gen and Lin Lin.
\newblock {G}enetic {A}lgorithms and {T}heir {A}pplications.
\newblock In {\em Springer Handbook of Engineering Statistics}, pages 635--674.
  Springer London, London, 2023.

\bibitem{RistoMik2024}
Risto Miikkulainen.
\newblock Generative {{AI}}: An {{AI}} paradigm shift in the making?
\newblock {\em AI Mag.}, 45:165--167, February 2024.

\bibitem{Bommasani2021}
Rishi Bommasani et~al.
\newblock On the {O}pportunities and {R}isks of {F}oundation {M}odels.
\newblock arXiv:2108.07258, 2022.

\bibitem{Yu2020}
Tong Yu and Hong Zhu.
\newblock Hyper-parameter optimization: A review of algorithms and
  applications.
\newblock arXiv:2003.05689, 2020.

\bibitem{Back2023}
Thomas H.~W. Bäck et~al.
\newblock Evolutionary {A}lgorithms for {P}arameter {O}ptimization—{T}hirty
  {Y}ears {L}ater.
\newblock {\em Evol. Comput.}, 31(2):81--122, June 2023.

\bibitem{Huang2020}
Changwu Huang, Yuanxiang Li, and Xin Yao.
\newblock A {S}urvey of {A}utomatic {P}arameter {T}uning {M}ethods for
  {M}etaheuristics.
\newblock {\em IEEE Trans. on Evol. Comput.}, 24(2):201--216, April 2020.

\bibitem{Hutter2019}
Frank Hutter, Lars Kotthoff, and Joaquin Vanschoren.
\newblock {\em Automated {M}achine {L}earning - {M}ethods, {S}ystems,
  {C}hallenges}.
\newblock Springer, 2019.

\bibitem{Martinez2021}
Aritz~D. Martinez et~al.
\newblock Lights and shadows in {E}volutionary {D}eep {L}earning: {T}axonomy,
  critical methodological analysis, cases of study, learned lessons,
  recommendations and challenges.
\newblock {\em Inf. Fusion}, 67:161--194, March 2021.

\bibitem{Real2020}
Esteban Real, Chen Liang, David So, and Quoc Le.
\newblock {A}uto{ML}-{Z}ero: Evolving {M}achine {L}earning {A}lgorithms {F}rom
  {S}cratch.
\newblock In {\em Proceedings of the 37th International Conference on Machine
  Learning}, volume 119, pages 8007--8019, July 2020.

\bibitem{Stanley201924}
Kenneth~O. Stanley, Jeff Clune, Joel Lehman, and Risto Miikkulainen.
\newblock Designing neural networks through neuroevolution.
\newblock {\em Nat. Mach. Intell.}, 1(1):24--35, January 2019.

\bibitem{Parisi2019}
German~I. Parisi et~al.
\newblock Continual lifelong learning with neural networks: {A} review.
\newblock {\em Neural Netw.}, 113:54–71, May 2019.

\bibitem{Stokel-Walker2023}
Chris Stokel-Walker and Richard Van~Noorden.
\newblock What {C}hat{GPT} and generative {AI} mean for science.
\newblock {\em Nat.}, 614(7947):214--216, February 2023.

\bibitem{Wang2019}
Rui Wang, Joel Lehman, Jeff Clune, and Kenneth~O. Stanley.
\newblock Paired {O}pen-{E}nded {T}railblazer {{(POET)}}: {E}ndlessly
  {G}enerating {I}ncreasingly {C}omplex and {D}iverse {L}earning {E}nvironments
  and {T}heir {S}olutions.
\newblock arXiv:1901.01753, 2019.

\bibitem{Wang2017}
Yu-Xiong Wang, Deva Ramanan, and Martial Hebert.
\newblock Learning to {M}odel the {T}ail.
\newblock In {\em Advances in Neural Information Processing Systems},
  volume~30, page 7032–7042, December 2017.

\bibitem{Wang2020}
Yaqing Wang, Quanming Yao, James~T. Kwok, and Lionel~M. Ni.
\newblock Generalizing from a {F}ew {E}xamples: {A} {S}urvey on {F}ew-{S}hot
  {L}earning.
\newblock {\em ACM Comput. Surv.}, 53(3):1--34, June 2020.

\bibitem{TangXin2021}
Ke~Tang, Shengcai Liu, Peng Yang, and Xin Yao.
\newblock Few-shots parallel algorithm portfolio construction via co-evolution.
\newblock {\em IEEE Trans. on Evol. Comput.}, 25(3):595--607, June 2021.

\bibitem{LiuXiao2023}
Xiao Liu et~al.
\newblock Self-{S}upervised {L}earning: {G}enerative or {C}ontrastive.
\newblock {\em IEEE Trans. on Knowl. and Data Eng.}, 35(1):857--876, January
  2023.

\bibitem{Fang2017}
Meng Fang, Jie Yin, Lawrence~O. Hall, and Dacheng Tao.
\newblock Active {M}ultitask {L}earning {W}ith {T}race {N}orm {R}egularization
  {B}ased on {E}xcess {R}isk.
\newblock {\em IEEE Trans. on Cybern.}, 47(11):3906–3915, November 2017.

\bibitem{Difrancesco2018}
Chiara {Di Francescomarino} et~al.
\newblock Genetic algorithms for hyperparameter optimization in predictive
  business process monitoring.
\newblock {\em Inf. Syst.}, 74:67--83, May 2018.

\bibitem{Darwish2020}
Ashraf Darwish, Aboul~Ella Hassanien, and Swagatam Das.
\newblock A survey of swarm and evolutionary computing approaches for deep
  learning.
\newblock {\em Artif. Intell. Rev.}, 53:1767--1812, June 2020.

\bibitem{Barthi2020}
Vandana Bharti, Bhaskar Biswas, and Kaushal~Kumar Shukla.
\newblock Recent {T}rends in {N}ature {I}nspired {C}omputation with
  {A}pplications to {D}eep {L}earning.
\newblock In {\em 10th International Conference on Cloud Computing, Data
  Science \& Engineering}, pages 294--299, January 2020.

\bibitem{Orive2014}
David Orive et~al.
\newblock Evolutionary algorithms for hyperparameter tuning on neural networks
  models.
\newblock In {\em Proceedings of the 26th european modeling \& simulation
  symposium}, pages 402--409, September 2014.

\bibitem{Miikkulainen2019}
Risto Miikkulainen et~al.
\newblock Evolving {D}eep {N}eural {N}etworks.
\newblock In {\em Artificial Intelligence in the Age of Neural Networks and
  Brain Computing (Second Edition)}, chapter~14, pages 269--287. Academic
  Press, 2024.

\bibitem{XueB2020}
Yanan Sun et~al.
\newblock Automatically {D}esigning {{CNN}} {A}rchitectures {U}sing the
  {G}enetic {A}lgorithm for {I}mage {C}lassification.
\newblock {\em IEEE Trans. on Cybern.}, 50(9):3840--3854, September 2020.

\bibitem{XueB2020b}
Yanan Sun, Bing Xue, Mengjie Zhang, and Gary~G. Yen.
\newblock Evolving {D}eep {C}onvolutional {N}eural {N}etworks for {I}mage
  {C}lassification.
\newblock {\em IEEE Trans. on Evol. Comput.}, 24(2):394--407, April 2020.

\bibitem{Yuqiao2023}
Yuqiao Liu et~al.
\newblock A {S}urvey on {E}volutionary {N}eural {A}rchitecture {S}earch.
\newblock {\em IEEE Trans. on Neural Netw. and Learn. Syst.}, 34(2):550--570,
  February 2023.

\bibitem{Espejo2010}
Pedro~G. Espejo, Sebastián Ventura, and Francisco Herrera.
\newblock A {S}urvey on the {A}pplication of {G}enetic {P}rogramming to
  {C}lassification.
\newblock {\em IEEE Trans. on Syst., Man, and Cybern., Part C (Appl. and
  Rev.)}, 40(2):121--144, March 2010.

\bibitem{Tran2016}
Binh Tran, Bing Xue, and Mengjie Zhang.
\newblock Genetic programming for feature construction and selection in
  classification on high-dimensional data.
\newblock {\em Memet. Comput.}, 8:3--15, March 2016.

\bibitem{lensen2019can}
Andrew Lensen, Bing Xue, and Mengjie Zhang.
\newblock Can genetic programming do manifold learning too?
\newblock In {\em Genetic Programming: 22nd European Conference}, pages
  114--130, April 2019.

\bibitem{Lensen2020}
Andrew Lensen, Mengjie Zhang, and Bing Xue.
\newblock Multi-objective genetic programming for manifold learning: balancing
  quality and dimensionality.
\newblock {\em Genet. Program. and Evol. Mach.}, 21(3):399--431, February 2020.

\bibitem{Elsken2019b}
Thomas Elsken, Jan~Hendrik Metzen, and Frank Hutter.
\newblock Neural {A}rchitecture {S}earch: A {S}urvey.
\newblock {\em J. of Mach. Learn. Res.}, 20(55):1--21, March 2019.

\bibitem{Zhou2021}
Xun Zhou, A.~K. Qin, Maoguo Gong, and Kay~Chen Tan.
\newblock A survey on {E}volutionary {C}onstruction of {D}eep {N}eural
  {N}etworks.
\newblock {\em IEEE Trans. on Evol. Comput.}, 25(5):894--912, October 2021.

\bibitem{Burke2009}
Edmund~K. Burke et~al.
\newblock Exploring {H}yper-heuristic {M}ethodologies with {G}enetic
  {P}rogramming.
\newblock In {\em Computational Intelligence: Collaboration, Fusion and
  Emergence}, pages 177--201. Springer Berlin Heidelberg, Berlin, Heidelberg,
  2009.

\bibitem{Burke2013}
Edmund~K Burke et~al.
\newblock Hyper-heuristics: A survey of the state of the art.
\newblock {\em J. of the Oper. Res. Soc.}, 64:1695--1724, July 2013.

\bibitem{Nguyen2012}
Trung~Thanh Nguyen, Shengxiang Yang, and Juergen Branke.
\newblock Evolutionary dynamic optimization: A survey of the state of the art.
\newblock {\em Swarm and Evol. Comput.}, 6:1--24, October 2012.

\bibitem{Azzouz2017}
Radhia Azzouz, Slim Bechikh, and Lamjed Ben~Said.
\newblock Dynamic {M}ulti-objective {O}ptimization {U}sing {E}volutionary
  {A}lgorithms: A {S}urvey.
\newblock In {\em Recent Advances in Evolutionary Multi-objective
  Optimization}, pages 31--70. Springer International Publishing, Cham, 2017.

\bibitem{Krawczyk2016}
Bartosz Krawczyk, Mikel Galar, Łukasz Jeleń, and Francisco Herrera.
\newblock Evolutionary undersampling boosting for imbalanced classification of
  breast cancer malignancy.
\newblock {\em Appl. Soft Comput.}, 38:714--726, January 2016.

\bibitem{Chen2023}
Shiming Chen, Shuhuang Chen, Wenjin Hou, Weiping Ding, and Xinge You.
\newblock {EGANS}: {E}volutionary {G}enerative {A}dversarial {N}etwork {S}earch
  for {Z}ero-{S}hot {L}earning.
\newblock {\em Early Access in IEEE Trans. on Evol. Comput.}, 2023.
\newblock to be published.

\bibitem{Tan2021}
Kay~Chen Tan, Liang Feng, and Min Jiang.
\newblock Evolutionary {T}ransfer {O}ptimization - {A} {N}ew {F}rontier in
  {E}volutionary {C}omputation {R}esearch.
\newblock {\em IEEE Comput. Intell. Mag.}, 16(1):22--33, February 2021.

\bibitem{Jiang2021}
Min Jiang et~al.
\newblock A {F}ast {D}ynamic {E}volutionary {M}ultiobjective {A}lgorithm via
  {M}anifold {T}ransfer {L}earning.
\newblock {\em IEEE Trans. on Cybern.}, 51(7):3417--3428, July 2021.

\bibitem{Wu2023}
Linjie Wu et~al.
\newblock Dynamic multi-objective evolutionary algorithm based on knowledge
  transfer.
\newblock {\em Inf. Sci.}, 636:118886, July 2023.

\bibitem{Reyes2018}
Oscar Reyes and Sebasti\'{a}n Ventura.
\newblock Evolutionary {S}trategy to {P}erform {B}atch-{M}ode {A}ctive
  {L}earning on {M}ulti-{L}abel {D}ata.
\newblock {\em ACM Trans. Intell. Syst. Technol.}, 9(4):1--26, January 2018.

\bibitem{Luo2022}
Rongjuan Luo, Shoufeng Ji, and Yuanyuan Ji.
\newblock An active-learning {P}areto evolutionary algorithm for parcel locker
  network design considering accessibility of customers.
\newblock {\em Comput. \& Oper. Res.}, 141:105677, May 2022.

\bibitem{Lely2014}
Daniel Le~Ly and Hod Lipson.
\newblock Optimal {E}xperiment {D}esign for {C}oevolutionary {A}ctive
  {L}earning.
\newblock {\em IEEE Trans. on Evol. Comput.}, 18(3):394--404, June 2014.

\bibitem{Osaba2022}
Eneko Osaba, Javier Del~Ser, Aritz~D Martinez, and Amir Hussain.
\newblock Evolutionary multitask optimization: a methodological overview,
  challenges, and future research directions.
\newblock {\em Cogn. Comput.}, 14(3):927--954, April 2022.

\bibitem{Xu2022}
Hao Xu, A.~K. Qin, and Siyu Xia.
\newblock Evolutionary {M}ultitask {O}ptimization {W}ith {A}daptive {K}nowledge
  {T}ransfer.
\newblock {\em IEEE Trans. on Evol. Comput.}, 26(2):290--303, April 2022.

\bibitem{Xiaoliang2019}
Xiaoliang Ma, Xiaodong Li, Qingfu Zhang, Ke~Tang, Zhengping Liang, Weixin Xie,
  and Zexuan Zhu.
\newblock A {S}urvey on {C}ooperative {C}o-{E}volutionary {A}lgorithms.
\newblock {\em IEEE Trans. on Evol. Comput.}, 23(3):421--441, June 2019.

\bibitem{Tan2018}
Chuanqi Tan et~al.
\newblock A {S}urvey on {D}eep {T}ransfer {L}earning.
\newblock In {\em Artificial Neural Networks and Machine Learning}, pages
  270--279, October 2018.

\bibitem{Stanley2002}
Kenneth~O. Stanley and Risto Miikkulainen.
\newblock Evolving {N}eural {N}etworks through {A}ugmenting {T}opologies.
\newblock {\em Evol. Comput.}, 10(2):99--127, June 2002.

\bibitem{Dufourq2017}
Emmanuel Dufourq and Bruce~A. Bassett.
\newblock {EDEN}: {E}volutionary deep networks for efficient machine learning.
\newblock In {\em Pattern Recognition Association of South Africa and Robotics
  and Mechatronics}, pages 110--115, November 2017.

\bibitem{Charte2020}
Francisco Charte, Antonio~J Rivera, Francisco Mart{\'\i}nez, and Mar{\'\i}a~J
  del Jesus.
\newblock {EvoAAA}: An evolutionary methodology for automated neural
  autoencoder architecture search.
\newblock {\em Integr. Computer-Aided Eng.}, 27(3):211--231, May 2020.

\bibitem{Assunccao2019}
Filipe Assun{\c{c}}ao, Nuno Louren{\c{c}}o, Penousal Machado, and Bernardete
  Ribeiro.
\newblock {{DENSER}}: deep evolutionary network structured representation.
\newblock {\em Genet. Program. and Evol. Mach.}, 20:5--35, September 2019.

\bibitem{Real2017large}
Esteban Real et~al.
\newblock Large-scale evolution of image classifiers.
\newblock In {\em International Conference on Machine Learning}, volume~70,
  pages 2902--2911, August 2017.

\bibitem{Liang2019}
Jason Liang et~al.
\newblock Evolutionary {N}eural {A}uto{ML} for {D}eep {L}earning.
\newblock In {\em Proceedings of the Genetic and Evolutionary Computation
  Conference}, page 401–409, July 2019.

\bibitem{martinez2021adaptive}
Aritz~D Martinez, Javier Del~Ser, Eneko Osaba, and Francisco Herrera.
\newblock Adaptive {M}ultifactorial {E}volutionary {O}ptimization for
  {M}ultitask {R}einforcement {L}earning.
\newblock {\em IEEE Trans. on Evol. Comput.}, 26(2):233--247, April 2021.

\bibitem{Lu2019}
Zhichao Lu et~al.
\newblock {{NSGA-Net}}: {N}eural {A}rchitecture {S}earch {U}sing
  {M}ulti-{O}bjective {G}enetic {A}lgorithm.
\newblock In {\em Proceedings of the Genetic and Evolutionary Computation
  Conference}, page 419–427, July 2019.

\bibitem{LuZhichao2020}
Zhichao Lu et~al.
\newblock {NSGANetV2}: {E}volutionary {M}ulti-objective {S}urrogate-{A}ssisted
  {N}eural {A}rchitecture {S}earch.
\newblock In {\em European Conference on Computer Vision}, pages 35--51, August
  2020.

\bibitem{LuZhichao2021}
Zhichao Lu et~al.
\newblock Neural {A}rchitecture {T}ransfer.
\newblock {\em IEEE Trans. on Patt. Anal. and Mach. Intell.}, 43(9):2971--2989,
  September 2021.

\bibitem{Elsken2019}
Thomas Elsken, Jan~Hendrik Metzen, and Frank Hutter.
\newblock Efficient {M}ulti-{O}bjective {N}eural {A}rchitecture {S}earch via
  {L}amarckian {E}volution.
\newblock In {\em International Conference on Learning Representations}, pages
  1--23, May 2019.

\bibitem{Phan2023}
Quan~Minh Phan and Ngoc~Hoang Luong.
\newblock Enhancing multi-objective evolutionary neural architecture search
  with training-free {P}areto local search.
\newblock {\em Appl. Intell.}, 53(8):8654--8672, August 2023.

\bibitem{Guha2023}
Ritam Guha et~al.
\newblock {MOAZ}: {A} {M}ulti-{O}bjective {A}uto{ML}-{Z}ero {F}ramework.
\newblock In {\em Proceedings of the Genetic and Evolutionary Computation
  Conference}, page 485–492, July 2023.

\bibitem{JasonMa2023}
Yecheng~Jason Ma et~al.
\newblock Eureka: {H}uman-{L}evel {R}eward {D}esign via {C}oding {L}arge
  {L}anguage {M}odels.
\newblock arXiv:2310.12931, 2023.

\bibitem{Wistuba2020}
Martin Wistuba.
\newblock {{XferNAS}}: {T}ransfer {N}eural {A}rchitecture {S}earch.
\newblock In {\em Proceedings of Machine Learning and Knowledge Discovery in
  Databases: European Conference}, page 247–262, September 2020.

\bibitem{Zhao2023}
Hong Zhao et~al.
\newblock What makes evolutionary multi-task optimization better: A
  comprehensive survey.
\newblock {\em Appl. Soft Comput.}, 145:110545, September 2023.

\bibitem{Wu2024}
Xingyu Wu et~al.
\newblock Evolutionary {C}omputation in the {E}ra of {L}arge {L}anguage
  {M}odel: {S}urvey and {R}oadmap.
\newblock arXiv:2401.10034, 2024.

\bibitem{Hemberg2024}
Erik Hemberg, Stephen Moskal, and Una-May O'Reilly.
\newblock Evolving {C}ode with {A} {L}arge {L}anguage {M}odel.
\newblock arXiv:2401.07102, 2024.

\bibitem{Guo2024}
Qingyan Guo et~al.
\newblock Connecting {L}arge {L}anguage {M}odels with {E}volutionary
  {A}lgorithms {Y}ields {P}owerful {P}rompt {O}ptimizers.
\newblock In {\em The Twelfth International Conference on Learning
  Representations}, May 2024.

\bibitem{Liu2023b}
Fei Liu, Xialiang Tong, Mingxuan Yuan, and Qingfu Zhang.
\newblock Algorithm {E}volution {U}sing {L}arge {L}anguage {M}odel.
\newblock arXiv:2311.15249, 2023.

\bibitem{Zhuang2021}
Fuzhen Zhuang et~al.
\newblock A {C}omprehensive {S}urvey on {T}ransfer {L}earning.
\newblock {\em Proceedings of the IEEE}, 109(1):43--76, January 2021.

\bibitem{Poyatos2023}
Javier Poyatos et~al.
\newblock Multiobjective evolutionary pruning of {D}eep {N}eural {N}etworks
  with {T}ransfer {L}earning for improving their performance and robustness.
\newblock {\em Appl. Soft Comput.}, 147:110757, November 2023.

\bibitem{Bali2020}
Kavitesh~Kumar Bali, Yew-Soon Ong, Abhishek Gupta, and Puay~Siew Tan.
\newblock Multifactorial {E}volutionary {A}lgorithm {W}ith {O}nline {T}ransfer
  {P}arameter {E}stimation: {{MFEA-II}}.
\newblock {\em IEEE Trans. on Evol. Comput.}, 24(1):69--83, February 2020.

\bibitem{Liu2023}
Songbai Liu et~al.
\newblock Evolutionary {M}ultitasking for {L}arge-{S}cale {M}ultiobjective
  {O}ptimization.
\newblock {\em IEEE Trans. on Evol. Comput.}, 27(4):863--877, August 2023.

\bibitem{XueB2020c}
Yanan Sun et~al.
\newblock Surrogate-{A}ssisted {E}volutionary {D}eep {L}earning {U}sing an
  {E}nd-to-{E}nd {R}andom {F}orest-{B}ased {P}erformance {P}redictor.
\newblock {\em IEEE Trans. on Evol. Comput.}, 24(2):350--364, April 2020.

\bibitem{Wang2022}
Xilu Wang, Yaochu Jin, Sebastian Schmitt, and Markus Olhofer.
\newblock Transfer {L}earning {B}ased {C}o-{S}urrogate {A}ssisted
  {E}volutionary {B}i-{O}bjective {O}ptimization for {O}bjectives with
  {N}on-{U}niform {E}valuation times.
\newblock {\em Evol. Comput.}, 30(2):221--251, June 2022.

\bibitem{correia2023evolutionary}
Jo{\~a}o Correia, Francisco Baeta, and Tiago Martins.
\newblock Evolutionary {G}enerative {M}odels.
\newblock In {\em Handbook of Evolutionary Machine Learning}, pages 283--329.
  Springer Nature Singapore, Singapore, 2024.

\bibitem{Taylor2019}
Tim Taylor.
\newblock Evolutionary {I}nnovations and {W}here to {F}ind {T}hem: {R}outes to
  {O}pen-{E}nded {E}volution in {N}atural and {A}rtificial {S}ystems.
\newblock {\em Artif. Life}, 25(2):207--224, May 2019.

\bibitem{Lehman2011}
Joel Lehman and Kenneth~O. Stanley.
\newblock Abandoning {O}bjectives: {E}volution {T}hrough the {S}earch for
  {N}ovelty {A}lone.
\newblock {\em Evol. Comput.}, 19(2):189--223, June 2011.

\bibitem{Brant2017}
Jonathan~C. Brant and Kenneth~O. Stanley.
\newblock Minimal {C}riterion {C}oevolution: {A} {N}ew {A}pproach to
  {O}pen-{E}nded {S}earch.
\newblock In {\em Proceedings of the Genetic and Evolutionary Computation
  Conference}, page 67–74, July 2017.

\bibitem{Neumann2019}
Aneta Neumann, Wanru Gao, Markus Wagner, and Frank Neumann.
\newblock Evolutionary {D}iversity {O}ptimization {U}sing {M}ulti-{O}bjective
  {I}ndicators.
\newblock In {\em Proceedings of the Genetic and Evolutionary Computation
  Conference}, page 837–845, July 2019.

\bibitem{preuss2015multimodal}
Mike Preuss.
\newblock {\em Multimodal optimization by means of evolutionary algorithms}.
\newblock Springer, 2015.

\bibitem{Tanabe2020}
Ryoji Tanabe and Hisao Ishibuchi.
\newblock A {R}eview of {E}volutionary {M}ultimodal {M}ultiobjective
  {O}ptimization.
\newblock {\em IEEE Trans. on Evol. Comput.}, 24(1):193--200, February 2020.

\bibitem{lehman2011novelty}
Joel Lehman and Kenneth~O. Stanley.
\newblock Novelty {S}earch and the {P}roblem with {O}bjectives.
\newblock In {\em Genetic Programming Theory and Practice IX}, pages 37--56.
  Springer New York, New York, 2011.

\bibitem{Langford2019}
Michael~Austin Langford, Glen~A. Simon, Philip~K. McKinley, and Betty H.~C.
  Cheng.
\newblock Applying {E}volution and {N}ovelty {S}earch to {E}nhance the
  {R}esilience of {A}utonomous {S}ystems.
\newblock In {\em IEEE/ACM 14th International Symposium on Software Engineering
  for Adaptive and Self-Managing Systems}, pages 63--69, May 2019.

\bibitem{ZhangQingquan2023}
Qingquan Zhang et~al.
\newblock Mitigating {U}nfairness via {E}volutionary {M}ultiobjective
  {E}nsemble {L}earning.
\newblock {\em IEEE Trans. on Evol. Comput.}, 27(4):848--862, August 2023.

\bibitem{WangXin2024}
Ziming Wang, Changwu Huang, Yun Li, and Xin Yao.
\newblock Multi-objective {F}eature {A}ttribution {E}xplanation {F}or
  {E}xplainable {M}achine {L}earning.
\newblock {\em ACM Trans. Evol. Learn. Optim.}, 4(1):1--32, feb 2024.

\bibitem{bai2023evolutionary}
Hui Bai, Ran Cheng, and Yaochu Jin.
\newblock Evolutionary {R}einforcement {L}earning: {A} {S}urvey.
\newblock {\em Intell. Comput.}, 2:0025, May 2023.

\bibitem{omidvar2021review}
Mohammad~Nabi Omidvar, Xiaodong Li, and Xin Yao.
\newblock A {R}eview of {P}opulation-{B}ased {M}etaheuristics for
  {L}arge-{S}cale {B}lack-{B}ox {G}lobal {O}ptimization—part {I}.
\newblock {\em IEEE Trans. on Evol. Comput.}, 26(5):802--822, October 2022.

\bibitem{carmichael2021learning}
Zachariah Carmichael, Tim Moon, and Sam~Ade Jacobs.
\newblock Learning interpretable models through multi-objective neural
  architecture search.
\newblock arXiv:2112.08645, 2021.

\bibitem{liu2021multi}
Jia Liu and Yaochu Jin.
\newblock Multi-objective search of robust neural architectures against
  multiple types of adversarial attacks.
\newblock {\em Neurocomputing}, 453:73--84, September 2021.

\bibitem{cheng2023neural}
Zhi Cheng et~al.
\newblock Neural architecture search for wide spectrum adversarial robustness.
\newblock In {\em Proceedings of the AAAI Conference on Artificial
  Intelligence}, volume~37, pages 442--451, February 2023.

\bibitem{bai2021ood}
Haoyue Bai et~al.
\newblock {NAS-OoD}: Neural architecture search for out-of-distribution
  generalization.
\newblock In {\em Proceedings of the IEEE/CVF International Conference on
  Computer Vision}, pages 8320--8329, October 2021.

\bibitem{gambella2024flatnas}
Matteo Gambella, Fabrizio Pittorino, and Manuel Roveri.
\newblock {FlatNAS}: optimizing {F}latness in {N}eural {A}rchitecture {S}earch
  for {O}ut-of-{D}istribution {R}obustness.
\newblock arXiv:2402.19102, 2024.

\bibitem{Qu2024integration}
Youzhi Qu et~al.
\newblock Integration of cognitive tasks into artificial general intelligence
  test for large models.
\newblock {\em Iscience}, 27(4), March 2024.

\bibitem{Zhang2023one}
Chaoning Zhang et~al.
\newblock One {S}mall {S}tep for {G}enerative {{AI}}, {O}ne {G}iant {L}eap for
  {{AGI}}: {A} {C}omplete {S}urvey on {C}hat{{GPT}} in {{AIGC}} {E}ra.
\newblock arXiv:2304.06488, 2023.

\end{thebibliography}

\end{document}